\pdfoutput=1
\documentclass[11pt]{article}

\usepackage{ACL2023}
\usepackage{array}
\usepackage[T1]{fontenc}
\usepackage{inconsolata}
\usepackage[utf8]{inputenc}
\usepackage{latexsym}
\usepackage{microtype}
\usepackage{times}

\usepackage{booktabs}
\usepackage{enumitem}
\usepackage{graphicx}
\usepackage{tabularx}
\usepackage{xcolor}
\usepackage{xspace}
\usepackage{colortbl}
\usepackage{caption}
\usepackage{subcaption}
\usepackage{comment}
\usepackage{tablefootnote}

\newcommand{\dataset}{BUMP\xspace}

\newif\ifcomments
\commentstrue
\ifcomments
\newcommand{\draftcomment}[3]{{\color{#2}[\textsc{#1} #3]}}
\else
\newcommand{\draftcomment}[3]{}
\fi

\newif\ifdetailed
\ifdetailed
\newcommand{\detailed}[1]{#1}
\newcommand{\abbreviated}[1]{}
\else
\newcommand{\detailed}[1]{}
\newcommand{\abbreviated}[1]{#1}
\fi

\newcommand{\rob}[1]{\draftcomment{Rob:}{purple}{#1}}

\newcommand{\joel}[1]{\draftcomment{Joel:}{olive}{#1}}

\newcommand{\sbt}{\,\begin{picture}(-1,1)(-1,-3)\circle*{3}\end{picture}\ }

\title{
\dataset: A Benchmark of Unfaithful Minimal Pairs for Meta-Evaluation of Faithfulness Metrics
}

\author{   
    Liang Ma$^1$ \quad
    Shuyang Cao$^2$ \quad 
    Robert L. Logan IV$^1$ \quad
    Di Lu$^1$ \\ \bf    
    Shihao Ran$^1$ \quad
    Ke Zhang$^1$ \quad      
    Joel Tetreault$^1$ \quad
    Alejandro Jaimes$^1$ \\
    $^1$Dataminr Inc. \quad $^2$University of Michigan, Ann Arbor \\
    \texttt{\{lma,rlogan,dlu,sran,kzhang,jtetreault,}\\
\texttt{ajaimes\}@dataminr.com} \quad \texttt{caoshuy@umich.edu}
}

\begin{document}
\maketitle
% \aoife{The abstract gets rather detailed towards the end, do we need all the detail at this point? Since the scores are only mentioned for one piece of the results analysis, I might leave it out.}
% \aoife{It's never mentioned in the paper whether this dataset will be released. Will it be?}\lm{added back}
\begin{abstract}
    The proliferation of automatic faithfulness metrics for summarization has produced a need for benchmarks to evaluate them.
    While existing benchmarks measure the correlation with human judgements of faithfulness on model-generated summaries, they are insufficient for diagnosing whether metrics are: 1) \emph{consistent}, i.e., indicate lower faithfulness as errors are introduced into a summary, 2) effective on \emph{human-written} texts, and 3) sensitive to different \emph{error types} (as summaries can contain multiple errors).
    % However, unfaithful content could also be generated by humans with substantially different unfaithfulness distributions, resulting in biased or incorrect conclusions of faithfulness evaluation metrics. 
    To address these needs, we present a benchmark of unfaithful minimal pairs (\dataset), a dataset of 889 \emph{human-written}, \emph{minimally different} summary pairs, where a single error %(from an ontology of 7 types) 
    is introduced to a summary from the CNN/DailyMail dataset to produce an unfaithful summary. %, each consisting of a faithful and an unfaithful summary.
    %\rob{This is too much info for the abstract in my opinion:    Specifically, given a faithful summary, it is edited by a human such that the edited summary mostly overlaps with the reference summary, but exhibits an unfaithful error, thus forming a required faithful/unfaithful summary pair. }
    %exhibit a specific type of error (e.g., hallucinated entities, events, coreference issues, etc.).
    We find \dataset complements existing benchmarks in a number of ways\detailed{.}\abbreviated{:}%
    \detailed{
    First, the summaries in \dataset are harder to discriminate (ROC AUC scores in the 50--70$\%$s vs. 80--90$\%$s) and less probable under SOTA summarization models.
    Second, \dataset enables measuring the consistency of metrics, and reveals that the most discriminative metrics tend not to be the most consistent.
    Finally, \dataset enables the measurement of metrics' abilities to handle different error types, and shows that coreference and predicate errors are particular areas of weakness for current metrics.
    }%
    \abbreviated{
    1) the summaries in \dataset are harder to discriminate and less probable under SOTA summarization models,
    2) unlike non-pair-based datasets, \dataset can be used to measure the consistency of metrics, and reveals that the most discriminative metrics tend not to be the most consistent, and
    3) unlike datasets containing generated summaries with multiple errors, \dataset enables the measurement of metrics' performance on individual error types.
    }

    % both in terms of the probability of unfaithful summaries under a summarization
    % \rob{
    %     Need to be cleaned up:
    %     - Evidence this dataset supplements existing benchmarks (what do we learn that isn't already known?)
    %         - Ability to handle different error types.
    %         - Argument our dataset is more challenging.
    %             ROC AUC scores between 50-70 vs. 80-90 in TRUE.
    %         - Insight that the most consistent metrics aren't always the most discriminative.
    % }
    % Similar to how linguistic minimal pairs are used to evaluate language models, we use \dataset to evaluate 
    % %abstractive summarization models and 
    % whether metrics can differentiate faithful from unfaithful summaries. %whether they assign lower scores to the human generated unfaithful summaries.  
    % %We also compare with the performance of these faithfulness metrics in datasets where unfaithful summaries are generated by models/rules,...
    % We find that BUMP is more challenging that existing datasets in that: 1) \rob{claim about ROC AUC}, and 2) 
    % Furthermore, we demonstrate that faithfulness metrics struggle to detect certain types of errors, and their capability rankings also differ from prior work that utilizes model generated data. 
    % \rob{I don't get the 2nd claim in the previous sentence.}
    % \rob{I feel like highlighting our result that the most consistent metrics are not the most discriminative could also be a good idea.}
    % \rob{Abstract needs a punchier ending.}
\end{abstract}
% \joel{the abstract is very good though a little long.  No need to change it now but something to flag if we go for a short paper.  That could almost be an intro all to itself.}

\section{Introduction}
%\aoife{It would be nice to list some specific research questions here}
%\rob{This intro is a bit long. Even though it is nice motivation, I cut the first paragraph since it is motivating faithfulness metrics in general instead of faithfulness metric evaluation.}

% With the proliferation of media sources around the globe and the need for quickly understanding the fast-paced world, people are frequently presented with summaries of critical news events or professional articles in specific areas. 

% Such unfaithful content in article summaries could be caused by unintentional human mistakes or adversarial human actions, such as for  political and economical purposes [XXX].
% Furthermore, due to recent advances in automatic text summarization systems that leverage large pre-trained models [XXX], it enables model generated summaries with human-level fluency [XXX]. 
% However, unfaithful summaries could also be generated by these systems in an implicit (e.g., training data defect or bias [XXX]) or a human guided way [XXX]. 

\begin{figure*}[!ht]
    \centering
    \includegraphics[width=0.95\textwidth]{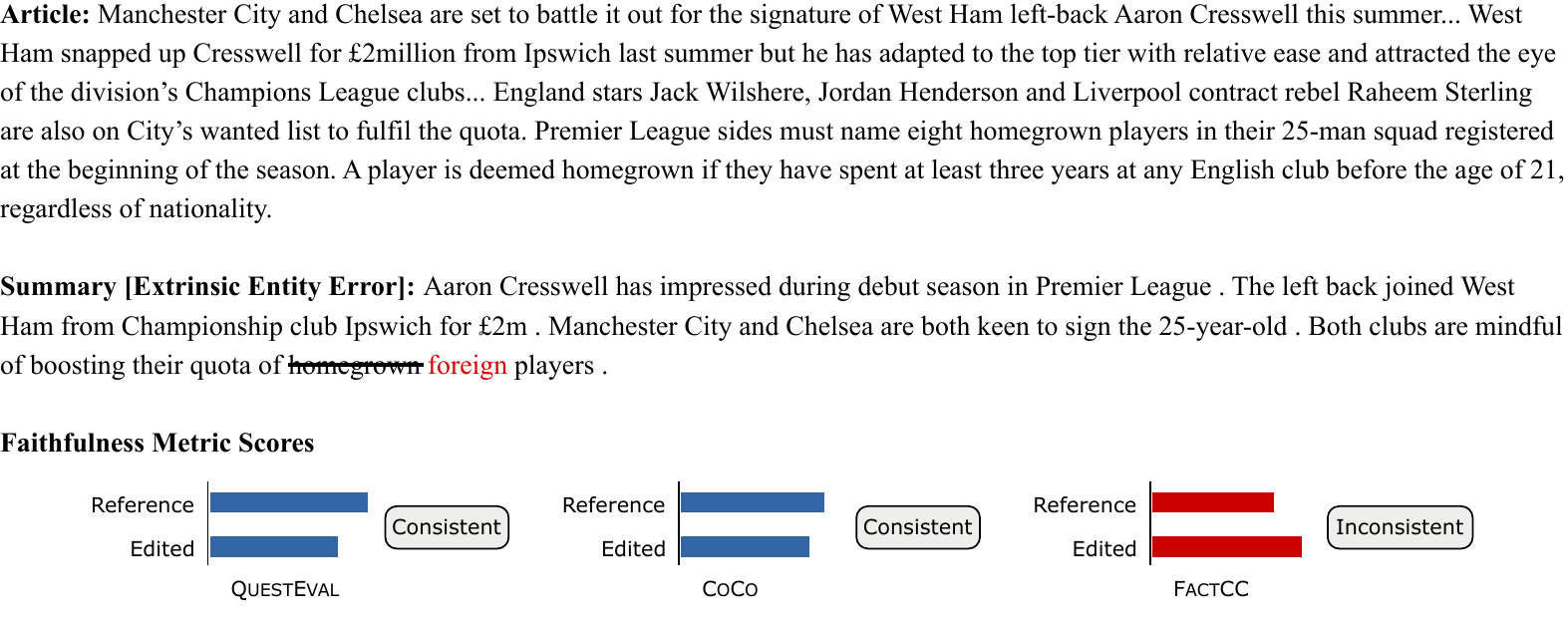}
    \caption{
        {\bf Example from \dataset dataset.}
        An annotator constructs an unfaithful summary containing an \textit{Extrinsic Entity Error} (Section \ref{subsec:taxonomy}) by replacing the word ``homegrown'' in the reference summary with the word ``foreign''.
        The reference and edited summary form a minimal unfaithful summary pair.
        Faithfulness metrics are evaluated on both the reference and edited summary and compared to measure whether the metric is consistent, e.g., in this example, \textsc{QuestEval} and \textsc{CoCo} are consistent, while \textsc{FactCC} is not.
    }
    \label{fig:hook}
\end{figure*}

%Our fast-paced world has promoted the spread of concise summaries instead of advocating original long articles.
%\rob{Previous sentence is a more of an opinion than a claim supported by evidence.}
%However, such summaries may be unfaithful to the original article, thus resulting in misinterpretation or even catastrophic decision making upon it, where 
% \emph{Unfaithful summary} of an article refers to anything that cannot be verified by or contradicts the original article; it may result in misinterpretation or even catastrophic decision making upon it.
Although modern abstractive summarization systems have improved 
% significantly
% \rob{I recommend against using words like 'drastically' in claims like this; it is subjective what a drastic improvement is. If reviewers have different opinions claims like this can put them in the mindset to disagree with you.} 
in their ability to produce fluent text~\cite{lewis-etal-2020-bart}, their ability to generate text that is factually grounded in the source article remains an issue~\cite{kryscinski-etal-2020-evaluating}.
This phenomenon has inspired the NLP community to develop faithfulness evaluation metrics \citep{fabbri-etal-2022-qafacteval, laban-etal-2022-summac, honovich-etal-2021-q2, scialom-etal-2021-questeval}
that automatically measure the extent to which abstractive summarization systems 
produce summaries that contain information that cannot be verified by the source article.
%and the source article. 

% However, such faithfulness scores are \emph{not} always reliable in capturing different types of unfaithful errors.
As the number of these automatic faithfulness metrics has increased, there has arisen a corresponding need for benchmarks that evaluate their relative strengths.
To satisfy this need, researchers have developed datasets such as FRANK \citep{pagnoni-etal-2021-understanding} and TRUE \citep{honovich-etal-2022-true} that are comprised of model-generated summaries along with human-annotated faithfulness levels.
% However, all unfaithful summaries in these datasets are model generated; 
% therefore, the evaluation results of these faithfulness metrics are \emph{only} valid in such model generated unfaithful content. 
% Yet, 
Although these datasets are useful for evaluating the degree to which faithfulness metrics correlate with human judgements and can discriminate unfaithful summaries, a number of factors limit the conclusions that can be drawn from them.
% \rob{TODO: Recap and expand on arguments from abstract here. The human-written, and error types parts are covered below, but we also should explain consistency.}
%\aoife{I might mention the Gabriel et al. paper here as the source of the consistency measurement.}\lm{added}
For one, because model summaries can vary in terms of length, content, and number of errors, these benchmarks are ill-suited for drawing conclusions about the \emph{consistency} \citep{gabriel-etal-2021-go} of metrics, i.e., whether their scores indicate lower faithfulness
% \joel{am going to be picky here: what decreases exactly?  it's not the metrics, it's the scores produced by the metrics?}
as summaries become increasingly unfaithful, as well as their sensitivity to specific \emph{types of errors}
since summaries can contain multiple errors.
Furthermore, as the summaries are machine-generated, these benchmarks cannot evaluate whether metrics can detect \emph{human-written} unfaithful summaries.
%\rob{I like this framing, however I think the minimal pair perspective should also be brought up in this paragraph since it is also part of the motivation for collecting \dataset}
% Additionally, the variation in the metric scores of summaries annotated with different error types by existing benchmarks may not accurately reflect the differences in the metric's ability to identify these errors since the content of the annotated summaries may contribute to the score difference.
% Furthermore, these unfaithful summaries \shuyang{here "these" is ambiguous} may differ from reference summaries substantially regarding text lengths, grammatical errors and multiple hallucinated words; therefore, evaluation conclusions of faithfulness metrics based on such noisy datasets become even less trustworthy.
%\rob{The end of this paragraph is meant to act as a sort of list of research questions---is this sufficient?}

To enable research on these topics, we present \dataset---a benchmark of unfaithful minimal pairs---a dataset of 889 minimally different summary pairs where all unfaithful summaries are generated by \emph{human annotators}.
% \footnote{Our benchmark will be released upon publication.}
% while maintaining the minimal difference comparing with faithful reference summaries, i.e., \emph{minimal pairs}. 
% In particular,
% \dataset is constructed from the CNN/DailyMail dataset \citep{hermann2015teaching}.
%\footnote{The reference summaries in CNN/DM are generally more faithful than other datasets, such as XSUM [XXX].}
% In particular, 
As illustrated in Figure~\ref{fig:hook}, given an article and its reference summary, we ask a human annotator to edit the reference summary in a minimal way such that the edited summary exhibits one unfaithful error.
% , while the difference between the edited and the reference summary (in terms of word overlapping) is as small as possible; accordingly, an unfaithful minimal pair is formed. 
%\rob{This is too much info for an intro:
%To build such dataset, we further perform two tasks: 
%(1) Taxonomy-based edits: The edited summary is required to exhibit one of the seven unfaithfulness error types in our taxonomy. 
%(2) Free-style edits: Annotators are not revealed with any unfaithfulness error types; they can freely edit reference summaries. 
%For Task~1, our taxonomy is defined by considering all pros and cons of taxonomies in \citet{pagnoni-etal-2021-understanding,tang-etal-2022-understanding}. 
%In addition, annotators in Task~2 are different from those in Task~1, thus avoiding potential error bias of edited summaries in Task~2. 
%}
We design two tasks for performance comparisons: 1) taxonomy-based edits, where a specific unfaithfulness error type is required according to our proposed taxonomy, and 
2) freestyle edits, where no error type constraints are imposed.
% In this way, we evaluate if faithfulness metrics experience performance variations in datasets generated in two different ways.
% \rob{This is not a particularly strong justification since it does not really describe why we would expect variation across the two settings. From the previous paragraph it is clear why taxonomy-based edits are important. However, it is not clear why free-edits are important.}
The motivation behind the first task setting is to ensure that different error types are adequately represented in our dataset, while the second task setting is important for understanding the completeness of our error type taxonomy as well as whether annotation difficulty is affected by instructing annotators to focus on specific error types.

We use \dataset to study the ability and performance consistency of faithfulness evaluation metrics in differentiating unfaithful summaries from faithful ones. 
%The reason we limit edits to minimal edits is that in prior works \citep{pagnoni-etal-2021-understanding,honovich-etal-2022-true}, model generated unfaithful datasets are uncontrollable and noisy ; 
%therefore, multiple reasons, e.g., text length, styles, grammatical errors, could contribute to poor faithfulness metric performance. 
Similar to how minimal pairs are used to diagnose linguistic knowledge of language models~\cite{marvin-linzen-2018-targeted, warstadt-etal-2020-blimp-benchmark}, the minimal summary pairs in \dataset allow targeted tests of a metric's consistency on different types of errors (Table~\ref{tab:taxonomy}).
This setup minimizes the effect of confounding factors that affect similar analyses (e.g., \citet{pagnoni-etal-2021-understanding} and \citet{tang-etal-2022-understanding}) such as text length, stylistic variation, and multiple errors occurring in the same summary.
% In addition, the small difference between faithful and unfaithful summaries in \dataset is potentially very subtle for the faithfulness metrics to detect, thus allowing us to test the robustness and reliability of these metrics via such a clean yet challenging dataset.
% \rob{The previous sentence includes a lot of vague claims that reviewers can contest just by disagreeing, e.g., that the differences are "very subtle" or that the dataset is "clean".}
We evaluate standard and state-of-the-art faithfulness metrics on \dataset using meta-evaluation protocols that target two phenomena:
1) \emph{consistency}, i.e. the fraction of unfaithful summaries that receive a lower score than their corresponding faithful summaries, and 
2) \emph{discriminability}, i.e., the metric's ability to classify unfaithful vs. faithful summaries as measured by ROC AUC.
%\rob{The previous sentences should be condensed into the paragraph before it.}

Our results (Section~\ref{sec:eval_result}) yield a number of useful findings:
1) \emph{\dataset differs substantially from existing benchmarks}: the summaries in \dataset are harder to discriminate (ROC AUC scores between 50--70$\%$ vs. 70--84$\%$) and are less probable under SOTA summarization models.
2) \emph{Discriminability != consistency}: interestingly, the most consistent metrics (\textsc{BARTScore}, \textsc{CoCo}) tend to have poor discriminability.
3) \emph{Some error types are harder than others}: e.g., metrics seem to uniformly struggle with summaries containing \textit{Intrinsic Error}s.

In sum, our contributions are three-fold:
1) We build a benchmark of human-generated unfaithful minimal pairs (\dataset) for evaluating faithfulness metrics.
2) We show human-generated unfaithful errors are substantially different from and more challenging than model-generated ones.
3) We demonstrate how \dataset provides insights on both the consistency and discriminative ability of faithfulness metrics on different error types than prior evaluation benchmarks that complement insights from existing benchmarks.
The BUMP dataset is available at: \url{https://github.com/dataminr-ai/BUMP}.
%and their performance consistency w.r.t. \emph{human} errors.\joel{this links to my comment above about ordering. In the prior para, discussion of entities comes first, but here it is a third point after the model generated ones}
%\item It also has \emph{social impact} on guiding us in how to use these faithfulness metrics to identify unfaithful human generated content. \rob{???}
%, which will be a great contribution to the community.

\section{Related Work}

Standard evaluation metrics for text generation tasks, e.g., \textsc{BLEU} and \textsc{ROUGE}, do not correlate well with human judgements of factual alignment in summarization settings~\cite{kryscinski-etal-2019-neural, maynez-etal-2020-faithfulness}.
This has motivated the development of automated faithfulness metrics that quantify factual alignment through methods that either: use NLI to measure the entailment degree between the source article and summary~\cite{kryscinski-etal-2020-evaluating, goyal-durrett-2020-evaluating, laban-etal-2022-summac}, 
compare summary probabilities when relevant information is removed from the source~\cite{xie-etal-2021-factual-consistency}, or use question answering models to measure if questions derived from the source can be answered by the summary and vice versa~\cite{wang-etal-2020-asking, durmus-etal-2020-feqa, scialom-etal-2021-questeval}.

Existing faithfulness metric evaluations use one of two classes of benchmarks: 1) machine-generated summaries paired with human-annotated faithfulness levels~\cite{laban-etal-2022-summac, pagnoni-etal-2021-understanding, tang-etal-2022-understanding}, and 2) summary pairs pertaining to the same article where one summary is faithful and the other is unfaithful~\cite{falke-etal-2019-ranking, gabriel-etal-2021-go}.
While both classes can evaluate a metric's ability to discriminate unfaithful summaries, the latter
%\ke{should be the former}
additionally allows for consistency tests, i.e., whether metrics assign higher values to more faithful summaries.

The \dataset dataset belongs to the second class of benchmarks; however, it has a number of unique properties.
First, unlike both \citet{falke-etal-2019-ranking} and \citet{gabriel-etal-2021-go}, the unfaithful summaries in \dataset are human-written.
In addition, the unfaithful summaries in \dataset are \emph{minimally different}, in the sense that only a single error differentiates the faithful and unfaithful summary.
As shown in Section~\ref{sec:eval_result}, this produces summary pairs that are substantially more challenging for metrics to differentiate.
% In addition, the unfaithful summaries in \dataset are \emph{minimally different}, in the sense that only a single error differentiates the faithful and unfaithful summary.
Inspired by the use of minimal pairs to diagnose linguistic knowledge of language models~\cite{marvin-linzen-2018-targeted, warstadt-etal-2020-blimp-benchmark}, the benefit of this approach is that it allows targeted tests of a metric's consistency on different types of errors (Section \ref{subsec:taxonomy}) while minimizing the effect of confounding factors.
Therefore, unlike other benchmarks with error type annotations~\cite{pagnoni-etal-2021-understanding,tang-etal-2022-understanding}, results on \dataset are not complicated by issues such as multiple errors appearing in the same summary.
% \ke{These should also goes to introduction part}

% FRANK~\cite{pagnoni-etal-2021-understanding},
% Go Figure~\cite{gabriel-etal-2021-go},
% Understanding Factual Errors~\cite{tang-etal-2022-understanding}

% \paragraph{Minimal Pairs}
% \cite{warstadt-etal-2020-blimp-benchmark, chomsky1965aspects, schutze1996empirical}

% Note that our reference and minimally edited summary pair is similar to the concept of linguistic minimal pairs in \citet{warstadt-etal-2020-blimp-benchmark}; 
% however, minimal pairs in \citet{warstadt-etal-2020-blimp-benchmark} focus on linguistic grammatical errors, while we explicitly ask annotators \emph{not} to generate edited summaries with grammatical errors (see Section~XXX for details). 
% Therefore, we ensure that our minimal pair only contains unfaithfulness errors.

% Datasets used for factural consistency evaluations. The evaluation protocol TRUE~\cite{honovich-etal-2022-true} consolidated 11 datasets acorss different tasks and domains. We focus on summarization and news domains? (although can be easily extend to other domains and tasks). Discuss the key difference and our novelty on minimum-difference pairs using human annotation vs model generated ones. TRUE claim NLI based method work well, while might not work well for our datasets. 

% How datasets are collected? What methods can be applied for evalaution (e.g., correlation analysis, binanay decision, or directional analysis, etc)
% \input{sections/30-Metrics.tex}
\section{Benchmark of Unfaithful Minimal Pairs (\dataset)}

Two annotation tasks are designed for \dataset, where Task~1 is taxonomy-based (a specific error type is required for the edited summary), and Task~2 allows freestyle edits (i.e., no error type constraints are imposed).
%\joel{should mention the two tasks here and not in the next sentence.  Since in 3.1 you do note that the two tasks}
In this section, we first describe how data sources are selected to build \dataset (\ref{sec:dataset}), and then describe the details of the two annotation tasks (\ref{subsec:taxonomy} and \ref{subsec:free-style}).

\subsection{Dataset}
\label{sec:dataset}

For Task~1, we randomly select $100$ article-summary pairs from the test set of the CNN/DailyMail dataset~\cite{hermann2015teaching}.%
 \footnote{
% While previous studies have annotated sample from the XSum dataset~\cite{narayan-etal-2018-dont}, the reference summaries in the XSum dataset frequently contain unfaithful content~\cite{maynez-etal-2020-faithfulness}, which is not ideal for our data collection purpose.
We do not annotate samples from the XSum dataset~\cite{narayan-etal-2018-dont} since the reference summaries are frequently unfaithful~\cite{maynez-etal-2020-faithfulness}.
}
For Task~2, we select an additional $100$ random article-summary pairs. %(i.e., $200$ in total.
Both tasks are performed via Amazon Mechanical Turk.\footnote{\url{https://www.mturk.com/};  annotation guidelines and interfaces are detailed in Appendices~\ref{sec:appendix1} and \ref{sec:appendix2}.}

\subsection{Task 1: Taxonomy-based Unfaithful Summaries}
\label{subsec:taxonomy}

% \rob{The structure of this paragraph is a little backwards in my opinion. Instead of starting with \emph{how} we collected annotations for task 1, start with a motivation for \emph{why} a taxonomy-based approach is important. This would naturally lead into the discussion for why we modified existing taxonomies. Only after that discuss how we collected this data.}
To obtain fine-grained performance evaluations of faithfulness metrics, it is critical to evaluate their sensitivity regarding various error types.
Furthermore, benchmarks should contain sufficiently many instances associated with each error type to enable statistically significant comparisons to be made.
To this end, we first define a taxonomy of unfaithful error types, and then ask annotators to introduce errors of a specific type in order to ensure each error type is adequately represented in the final dataset.
% \rob{Personal preference is not to use shorthand like this since it is informal}
% each error type in this taxonomy so as to control the data volume for each error type.

%The taxonomy for Task~1 is defined in Table~\ref{tab:taxonomy}.
%The reason to define our own taxonomy is that 
We note that existing taxonomies of error types may contain overlapped error types, e.g., grammatical vs. entity errors in FRANK~\citep{pagnoni-etal-2021-understanding} or lack fine granularity, e.g., \citet{tang-etal-2022-understanding}. By considering the strengths and shortcomings of existing taxonomies, we define our own taxonomy in Table~\ref{tab:taxonomy}.
%\joel{we need to say where this taxonomy came from.  Was this something we develoed on our own?  If so, what was the process?  If it is based on or related to prior work, we need to cite that here.  WAIT - it is on the next page....  } 
% are defined in Table~\ref{tab:taxonomy}.
Our taxonomy is first adapted from the one in FRANK~\citep{pagnoni-etal-2021-understanding} by including semantic frame errors (\textit{Predicate Error}, \textit{Entity Error}, and \textit{Circumstance Error}) and \textit{Coreference Error}, and removing error types that might overlap with others. % (e.g., grammatical error) and cause confusion in the annotation task.
To further categorize each semantic frame error, we adopt the notions of \textit{Intrinsic} and \textit{Extrinsic} errors~\cite{maynez-etal-2020-faithfulness, goyal-durrett-2020-evaluating, tang-etal-2022-understanding}.
Note that we do not simply categorize errors into the \textit{Intrinsic} and \textit{Extrinsic} ones, 
as we believe semantic frame errors can better instruct annotators to create summaries with diverse unfaithful errors.
%in our pilot study~\shuyang{can't remember if we have done this}.
In our taxonomy,
the \textit{Intrinsic}/\textit{Extrinsic} distinction only applies to the \textit{Predicate, Entity, and Circumstance Error}, since for a \textit{Coreference Error}, it is generally ambiguous whether an erroneous pronoun/reference that does not exist in the source article should be regarded as intrinsic or extrinsic.
In total, this results in seven different error types. 
%\joel{should write what motivates our taxonomy.  It seems that we are taking the union of existing ontologies, if so, we can make that explicit.  Right now there is no justification for doing what we are doing.  }

For each of the seven error types in this taxonomy, given an article-summary pair, we ask the annotator to introduce an error of the required type through a minimal edit to the reference summary.
All \emph{<article, summary, error type>} Human
Intelligence Tasks (HITs) in Amazon Mechanical Turk are shuffled and there is no annotation repetition, i.e., one assignment per HIT.
This increases the chance that edits of the same reference summary will be made by different annotators. 
Additional details regarding qualification tests and annotation instructions are presented in Appendix~\ref{sec:appendix1}. 
%\aoife{This is a little unclear. It's clear that there is one change per annotation. But what is not clear is how many annotators you recruited? Whether each annotator is asked to provide 7 edits for one article? Did you have any double-annotated data?}

After the data collection, we manually check the validity of each edit.
For cases where the edits do not match the required error types, we relabel them with the corrected error types based on our taxonomy. 
The dataset statistics after correction are shown in Table~\ref{tab:dataset-stats}.
For this task, one common mistake is that  annotators consider the quantity of a \emph{noun object} as a circumstance and make edits to the quantity (the first example in Table~\ref{tab:incorrect_errors}), hence mistakenly treat \textit{Entity Error}s as \textit{Circumstance Error}s, which causes the total number of \textit{Circumstance Error}s to be only 160 (much smaller than that of \textit{Entity Error}s; see Table~\ref{tab:dataset-stats}). Another frequent mistake is that the edited word actually exists in the original article for the required extrinsic error (the second example in Table~\ref{tab:incorrect_errors}), which results in a smaller number of \textit{Extrinsic Error}s than intrinsic ones across all semantic frame errors, especially for \textit{Predicate Error}s.
Furthermore, Table~\ref{tab:dataset-stats}  shows all edited summaries can be categorized by our taxonomy (no summaries are relabeled as ``Other''), and the incorrect response rate is $16\%$, suggesting that, in general, annotators correctly respond with the required error types.

\begin{table}[tb]
    \small
    \centering
    \begin{tabularx}{\columnwidth}{lX}
        \toprule
        \midrule
        Error Type & Description \\ % & Example \\
        \midrule
        \midrule
        Predicate Error
            & 
            The predicate in the summary is inconsistent with the source article.
            % &
            % The Ebola vaccine was produced by the FDA in 2019.
            \\
        \midrule
        Entity Error
            &
            The subject/object of a predicate is inconsistent with the source article.
            % & 
            % The COVID-19 vaccine was approved by the FDA in 2019.
            \\
        \midrule
        Circumstance Error &
            Time, duration, or location of an event of the predicate is wrong.
            % &
            % The first vaccine for Ebola was approved by the FDA in 2014.
            \\
        \midrule
        Coreference Error &
            A pronoun/reference with wrong or nonexistent antecedent.
            % &
            % The first vaccine for Ebola was approved in 2019. They say a vaccine for COVID-19 is unlikely to be ready this year.
            \\
        \midrule
        \midrule
        Intrinsic Error &
            Error derived from information within the source article.
            \\
        \midrule
        Extrinsic Error &
            Error contains information not present in the source article.
            \\
        \midrule
        \bottomrule
    \end{tabularx}
    \caption{{\bf Error type taxonomy.}}
    \label{tab:taxonomy}
\end{table}

\subsection{Task 2: Freestyle Unfaithful Summaries}
\label{subsec:free-style}

In addition to the taxonomy-based Task~1,
we also conduct a separate task, 
Task~2, where annotators can edit reference summaries in any way they want, 
i.e., freestyle editing, as long as only one error is introduced to the reference summary via minimal edits.
The goal of Task~2 is to understand how human-generated unfaithful summaries may vary, 
and how the performance of faithfulness evaluation metrics changes accordingly, 
when there are no error type constraints. 
In particular, only annotators who did not participate in the qualification test of Task~1 are considered to participate in this task;
%Between Tasks~1 and 2, there is no annotator overlap (annotators in one task are \emph{not} aware of the existence of the other task); 
in this way, we ensure the edited summaries in Task~2 are not constrained to any known error types. 

%\aoife{How do you ensure that annotators don't know about the other task? Is it done through qualification? There were two separate qualification tasks, so do you do it be annotator id? I think it would be clearer to highlight the fact that you ensured no annotator overlap through some AMT mechanism rather than by a more indirect assumption that they don't know about the other tasks.}\lm{added}

To post-process all data collected in Task~2, 
we manually assign an error type to each data point, 
based on our error type taxonomy in Task~1.
%Similar to Task~1, to assign error types to data in Task~2, we also allow ``Other'' error type if the unfaithful summary cannot be described by our taxonomy or contains more than one error.
Without informing annotators of any specific error types, we observe the rate that the ``Other'' label occurs is only $2.5\%$ for Task~2 in Table~\ref{tab:dataset-stats}. 
This confirms that the vast majority of errors produced by humans adhere to our proposed taxonomy.
% \joel{is efficiency the right word here?  maybe we just say that it confirms the comprehensiveness of our taxonomy....}
For more details on Task~2, please
see Appendix~\ref{sec:appendix2}.
%\aoife{You later mention the Other category for cases where the original scheme didn't fit. I would mention this briefly here too. Also, did you ever disagree on what error type it was? How difficult is it for experts to apply this annotation scheme?} \lm{added}

\begin{table}[tb]
    \centering
    \small
    \begin{tabular}{llcc}
        \toprule
        \midrule
                        &           & Task~1 & Task~2 \\
         \midrule
         \midrule
         Predicate      & Intrinsic & 116    & 17 \\
                        & Extrinsic & 76    & 28 \\
         Entity         & Intrinsic & 128    & 28 \\
                        & Extrinsic & 115    & 62 \\
         Circumstance   & Intrinsic & 82    & 22 \\
                        & Extrinsic & 78    & 33 \\
         Coreference    & -         & 98    & 1  \\
         Other          & -         & 0     & 5  \\
         \midrule
         Total          &           & 693   & 196 \\
         \midrule
         \bottomrule
    \end{tabular}
    \caption{
     \bf{\dataset dataset statistics.}
    % \tablefootnote{We aimed to collect 700 and 200 summary pairs for Task~1 and 2, respectively. However, after removing incomplete assignments and summaries that are obviously not minimally edited, the final number of minimally edited summary pairs in Table~\ref{tab:dataset-stats} is not the same as we intended initially.}
    }
    \label{tab:dataset-stats}
\end{table}

\smallskip
\noindent\textbf{Remark.}
%\joel{just a note that the first half of this (long) paragraph is discussed earler in the paper.  Though it is more filled out here.}\lm{removed something}
For both tasks, we ask annotators to introduce only \emph{one} error (by editing the reference summary in a minimal way). 
\begin{comment}
The advantages are 
(i) Intuitively, the more similarity there is between the reference summary and the unfaithful summary, the more difficult it will be for the faithfulness evaluation metric to determine which one is unfaithful, 
thus enabling us to test the ability of faithfulness evaluation metrics in this challenging scenario. 
(ii) Since only minimal edits are allowed, 
the resulting summaries are clean in the sense that other confounding factors (summary length, style, etc.) can be eliminated 
to reliably study the ability of faithfulness evaluation metrics to discern different error types.
\joel{I added a para break just to break it up}
Furthermore, 
\end{comment}
We acknowledge that some reference summaries may be unfaithful in the first place; 
nevertheless, for both tasks, edited summaries are based on reference summaries, 
by which we ensure the edited summaries are always more unfaithful than reference summaries.
%\ke{also we should circle back to this on how our meta-evaluation protocol (e.g., sensitivity) could help to address this.} 
%The statistics of datasets from these two tasks are shown in Table~\ref{tab:dataset-stats}. Note that for Task~1, the statistics in Table~\ref{tab:dataset-stats} are based on our corrected error types, thus are not evenly distributed across the seven error types as we intended initially. When we correct or assign error types for Task~1 or 2, we allow ``Other'' as the error type if the unfaithful summary cannot be described by our taxonomy or contains more than one error type; nevertheless, the rate of ``Other'' is very small, i.e., $2.5\%$ for Task~2 and $0\%$ for Task~1, confirming the efficiency of our taxonomy. 
% although we require seven error types for each article and reference summary pair.

\begin{table*}[tb]
    \scriptsize
    \centering
    \begin{tabular}{m{0.4\textwidth}m{0.175\textwidth}m{0.05\textwidth}m{0.175\textwidth}m{0.05\textwidth}m{0.05\textwidth}}
        \toprule
	\centering Article (Partial) & \centering Reference Summary & \centering Required Error Type & \centering Edited Summary & \centering Corrected Error Type \tabularnewline
 % & Analysis\\
 
% & High faithful score from the metrics\\
\midrule
... The drugs, whose value is estimated at more than \textbf{\$105 million}, ... Officers arrested one Venezuelan and two Spanish citizens who were on board the vessel off the coast ... French customs officials seized nearly \textbf{250} kilograms (550 pounds) of cocaine on a vessel that was also off the coast of Martinique, according to authorities.  & The value of the drugs is estimated at more than \textbf{\$105} million. Officers arrested one Venezuelan and two Spanish citizens on board the vessel. & Intrinsic Circumstance Error & The value of the drugs is estimated at more than \textbf{\$250} million . Officers arrested one Venezuelan and two Spanish citizens on board the vessel. & Intrinsic Entity Error \\
% & Mistaken Entity Errors as Circumstance Errors \\
\midrule
Lightning, floods and a deluge of hailstones descended on St Louis Tuesday as powerful storms \textbf{pummeled} the mid-United States. Roads around the Missouri city were flooded in the intense downpour, with one town recording more than two inches of rain in half an hour. ...& St Louis \textbf{was hit} Tuesday by flash floods. A nearby town had more than two inches of rain in less than half an hour.& Extrinsic Predicate Error & St Louis \textbf{pummeled} Tuesday by flash floods . A nearby town had more than two inches of rain in less than half an hour.& Intrinsic Predicate Error \\
% & Mistaken Intrinsic Errors as Extrinsic Errors \\
         \bottomrule
    \end{tabular}
    \caption{
        {\bf Example error type corrections.}
        The above examples illustrate instances where annotators' edits to the reference do not match the \emph{required error type} they are requested to produce.
        For such examples, \dataset includes a manually annotated \emph{corrected error type}.
    }
    \label{tab:incorrect_errors}
\end{table*}

\section{Meta-Evaluation of Faithfulness Evaluation Metrics}
\label{sec:eval}

In this section, we first describe the faithfulness evaluation metrics benchmarked on \dataset (\ref{sec:eval_setup}). Then meta-evaluation protocols are discussed (\ref{sec:meta_eval}).

\subsection{Faithfulness Metrics}
\label{sec:eval_setup}

% \joel{should say how many metrics we look at total, and how many in each category.  Can do the by-category ones can say it in the respect paras}

% \paragraph{Evaluation Metrics.}

To cover diverse types of faithfulness metrics, in this section, we select metrics that are generally used for measuring generation quality (i.e., $n$-gram-based metrics), recent metrics that are proposed specifically for faithfulness evaluations, as well as some pre-trained model based metrics, which are detailed as follows. We investigate their abilities to distinguish faithful summaries from their minimally edited counterparts.
% \footnote{For metric scores that are not in $[0,1]$ ($1$: highest faithfulness score), they are normalized to be in that range.}

%\rob{The following list provides way to much detail. I think it would be fine to just list the methods, and only provide details that aren't available in their respective papers.} \lm{agreed}

\smallskip
\noindent
$\sbt\ $\textbf{$n$-Gram-Based Metrics:} We evaluate the following 2 $n$-gram-based metrics: \textsc{BLEU} \citep{10.3115/1073083.1073135} and \textsc{ROUGE} (\textsc{ROUGE-2} Precision specifically) \citep{lin-2004-rouge}.

\smallskip
\noindent
$\sbt\ $\textbf{Faithfulness Evaluation Metrics:} We evaluate the following 7 faithfulness evaluation metrics: \textsc{QuestEval}~\cite{scialom-etal-2021-questeval}, \textsc{$Q^2$}~\cite{honovich-etal-2021-q2}, \textsc{QAFactEval}~\cite{fabbri-etal-2022-qafacteval}, \textsc{FactCC}~\cite{kryscinski-etal-2020-evaluating}, \textsc{DAE}~\cite{goyal-durrett-2020-evaluating}, \textsc{SummaC}~\cite{laban-etal-2022-summac} and \textsc{CoCo}~\cite{xie-etal-2021-factual-consistency} in this paper.
To obtain a score for \textsc{FactCC}, we take the classifier probability of the summary being faithful.

\smallskip
\noindent
$\sbt\ $\textbf{Other Metrics:} We evaluate the following 3 pre-trained model based metrics: \textsc{BLEURT} \citep{sellam-etal-2020-bleurt} with the \textsc{BLEURT}-20 checkpoint \citep{pu-etal-2021-learning}, \textsc{BERTScore}~\citep{Zhang*2020BERTScore:} (specifically the \textsc{BERTScore}-precision variant) using the DeBERTa-xl-MNLI model~\citep{he2021deberta}, and \textsc{BARTScore}~\citep{yuan2021BARTScore} with a BART~\citep{lewis-etal-2020-bart} model fine-tuned on the CNN/DailyMail dataset.

Note that for reference-based metrics, faithfulness scores are computed by treating the input article as the reference, and the reference/edited summary as the system output.
We also normalize the direction of the metric score so that a higher score always corresponds to better faithfulness from the metric's view, e.g., \textsc{FactCC} predicts the probability that a summary is unfaithful, and so to obtain a faithfulness score, we take its complement.

\subsection{Meta-Evaluation}
\label{sec:meta_eval}

% \joel{I'm not quite sure what the evaluation is.  I think we're just looking at these metrics on BUMP?  But are the correct summaries also included?  

% Also what is meant by "summary pair"?

% Basically, what is the input to the eval and the output?  I'm getting confused by this para}

% \aoife{This section is rather repetitive. A lot of space is spent saying the same thing about ROC AUC.}
% The uniqueness of \dataset is for each summary pair, it is guaranteed that one of them is unfaithful and written by a human. Therefore, 
Each faithfulness metric takes an article-summary pair and outputs a numerical faithfulness score.
In our analysis, we measure faithfulness scores for both the
reference summary as well as the human-annotated erroneous summary.
% Faitwhere the summary is either the original faithful reference summary or the unfaithful edited summary by human annotators.
We quantify the difference between faithfulness metrics on \dataset using two measurement protocols: {\bf consistency} and {\bf ROC AUC}.
% We first study how faithfulness evaluation metrics \emph{consistently} correlate with the unfaithful information in a summary.
% To this end, we report %a meta-evaluation protocol called
% \textbf{consistency} \citep{gabriel-etal-2021-go}, which is
Originally introduced in \citet{gabriel-etal-2021-go}, consistency measures the success rate of a metric assigning a lower faithfulness score to the erroneous unfaithful summary.
% This is similar to the consistency measurement introduced by \citet{gabriel-etal-2021-go} which studies the correlation between the faithfulness metric and the number of errors in a summary.
In contrast, 
ROC AUC instead measures the overall capability of a metric to discriminate faithful from unfaithful content for an input summary, and has previously been used by \citet{honovich-etal-2022-true} for meta-evaluation. % , i.e., by treating the faithfulness detection as a binary classification problem and using the faithfulness metric as a faithfulness score.
% Discriminability has been studied in numerous other faithfulness metric evaluations~\cite{laban-etal-2022-summac,honovich-etal-2022-true}
Although other metrics such as balanced accuracy have also been used to evaluate disciminability~\cite{laban-etal-2022-summac}, we opt to use ROC AUC as it does not require determining a decision threshold.

% This is formulated as a binary classification problem (i.e., faithful and unfaithful summaries are assigned binary labels respectively as in \citet{honovich-etal-2022-true}). 
% and the meta-evaluation protocol we utilize is \textbf{ROC AUC}.

\section{Results}
\label{sec:eval_result}
We report and analyze the performance of faithfulness metrics in this section using meta-evaluation protocol consistency and ROC AUC.

\begin{table*}[t]
    \scriptsize
    \begin{subtable}[t]{\linewidth}
    \centering
    \begin{tabular}{llccccccccccc}
        \toprule
	&	BART	&	{CoCo}	&	{DAE}	&	QAFaEv	&	BERT	&	QuEv	&	{BLEURT}	&	\textsc{SummaC}	&	R-2	&	{BLEU}	&	$Q^2$	&	\textsc{FactCC}	\\
\midrule
% Intrinsic Predicate	&	96.6	&	88.8	&	87.1	&	79.3	&	81.0	&	69.0	&	69.8	&	61.2	&	51.7	&	39.7	&	49.1	&	49.1	\\
% Extrinsic Predicate	&	94.7	&	93.4	&	84.2	&	86.8	&	88.2	&	73.7	&	73.7	&	67.1	&	63.2	&	61.8	&	59.2	&	60.5	\\
% Intrinsic Entity	&	93.0	&	91.4	&	88.3	&	91.4	&	80.5	&	82.8	&	76.6	&	75.8	&	65.6	&	60.9	&	75.0	&	61.7	\\
% Extrinsic Entity	&	97.4	&	96.5	&	93.9	&	90.4	&	88.7	&	87.0	&	90.4	&	79.1	&	86.1	&	87.0	&	80.9	&	59.1	\\
% Intrinsic Circumstance	&	85.4	&	84.2	&	86.6	&	81.7	&	67.1	&	74.4	&	63.4	&	74.4	&	51.2	&	53.7	&	68.3	&	63.4	\\
% Extrinsic Circumstance	&	85.9	&	85.9	&	85.9	&	85.9	&	79.5	&	78.2	&	75.6	&	73.1	&	79.5	&	74.4	&	68.0	&	59.0	\\
% Coreference Error	&	86.7	&	92.9	&	86.7	&	70.4	&	82.7	&	82.7	&	67.4	&	46.9	&	72.5	&	86.7	&	56.1	&	65.3	\\
Overall & {\cellcolor[HTML]{F5D949}} \color[HTML]{000000} 91.9 & {\cellcolor[HTML]{F8CF3A}} \color[HTML]{000000} 90.8 & {\cellcolor[HTML]{FBB61A}} \color[HTML]{000000} 87.9 & {\cellcolor[HTML]{FB9606}} \color[HTML]{000000} 84.0 & {\cellcolor[HTML]{F78212}} \color[HTML]{F1F1F1} 81.4 & {\cellcolor[HTML]{EF6C23}} \color[HTML]{F1F1F1} 78.6 & {\cellcolor[HTML]{DE5238}} \color[HTML]{F1F1F1} 74.5 & {\cellcolor[HTML]{BA3655}} \color[HTML]{F1F1F1} 68.4 & {\cellcolor[HTML]{B3325A}} \color[HTML]{F1F1F1} 67.2 & {\cellcolor[HTML]{AB2F5E}} \color[HTML]{F1F1F1} 66.1 & {\cellcolor[HTML]{A82E5F}} \color[HTML]{F1F1F1} 65.7 & {\cellcolor[HTML]{7C1D6D}} \color[HTML]{F1F1F1} 59.5 \\
\midrule
Intrinsic Predicate & {\cellcolor[HTML]{F8FB9A}} \color[HTML]{000000} 96.6** & {\cellcolor[HTML]{FBBE23}} \color[HTML]{000000} 88.8 & {\cellcolor[HTML]{FCB014}} \color[HTML]{000000} 87.1 & {\cellcolor[HTML]{F1711F}} \color[HTML]{F1F1F1} 79.3 & {\cellcolor[HTML]{F67E14}} \color[HTML]{F1F1F1} 81.0 & {\cellcolor[HTML]{BD3853}} \color[HTML]{F1F1F1} 69.0 & {\cellcolor[HTML]{C33B4F}} \color[HTML]{F1F1F1} 69.8 & {\cellcolor[HTML]{88226A}} \color[HTML]{F1F1F1} 61.2 & {\cellcolor[HTML]{450A69}} \color[HTML]{F1F1F1} 51.7 & {\cellcolor[HTML]{000004}} \color[HTML]{F1F1F1} 39.7 & {\cellcolor[HTML]{310A5C}} \color[HTML]{F1F1F1} 49.1 & {\cellcolor[HTML]{310A5C}} \color[HTML]{F1F1F1} 49.1 \\
Extrinsic Predicate & {\cellcolor[HTML]{F1F179}} \color[HTML]{000000} 94.7 & {\cellcolor[HTML]{F2E661}} \color[HTML]{000000} 93.4 & {\cellcolor[HTML]{FB9706}} \color[HTML]{000000} 84.2 & {\cellcolor[HTML]{FCAC11}} \color[HTML]{000000} 86.8 & {\cellcolor[HTML]{FBBA1F}} \color[HTML]{000000} 88.2 & {\cellcolor[HTML]{D94D3D}} \color[HTML]{F1F1F1} 73.7 & {\cellcolor[HTML]{D94D3D}} \color[HTML]{F1F1F1} 73.7 & {\cellcolor[HTML]{B1325A}} \color[HTML]{F1F1F1} 67.1 & {\cellcolor[HTML]{972766}} \color[HTML]{F1F1F1} 63.2 & {\cellcolor[HTML]{8D2369}} \color[HTML]{F1F1F1} 61.8 & {\cellcolor[HTML]{7A1D6D}} \color[HTML]{F1F1F1} 59.2 & {\cellcolor[HTML]{84206B}} \color[HTML]{F1F1F1} 60.5 \\
Intrinsic Entity & {\cellcolor[HTML]{F3E35A}} \color[HTML]{000000} 93.0 & {\cellcolor[HTML]{F6D543}} \color[HTML]{000000} 91.4 & {\cellcolor[HTML]{FBBA1F}} \color[HTML]{000000} 88.3 & {\cellcolor[HTML]{F6D543}} \color[HTML]{000000} 91.4 & {\cellcolor[HTML]{F57B17}} \color[HTML]{F1F1F1} 80.5 & {\cellcolor[HTML]{F98C0A}} \color[HTML]{F1F1F1} 82.8 & {\cellcolor[HTML]{E75E2E}} \color[HTML]{F1F1F1} 76.6 & {\cellcolor[HTML]{E45A31}} \color[HTML]{F1F1F1} 75.8 & {\cellcolor[HTML]{A62D60}} \color[HTML]{F1F1F1} 65.6 & {\cellcolor[HTML]{87216B}} \color[HTML]{F1F1F1} 60.9 & {\cellcolor[HTML]{E05536}} \color[HTML]{F1F1F1} 75.0 & {\cellcolor[HTML]{8C2369}} \color[HTML]{F1F1F1} 61.7 \\
Extrinsic Entity & {\cellcolor[HTML]{FCFFA4}} \color[HTML]{000000} 97.4 & {\cellcolor[HTML]{F8FB9A}} \color[HTML]{000000} 96.5 & {\cellcolor[HTML]{F2EA69}} \color[HTML]{000000} 93.9 & {\cellcolor[HTML]{F9CB35}} \color[HTML]{000000} 90.4 & {\cellcolor[HTML]{FBBE23}} \color[HTML]{000000} 88.7 & {\cellcolor[HTML]{FCAE12}} \color[HTML]{000000} 87.0 & {\cellcolor[HTML]{F9CB35}} \color[HTML]{000000} 90.4 & {\cellcolor[HTML]{F06F20}} \color[HTML]{F1F1F1} 79.1 & {\cellcolor[HTML]{FCA60C}} \color[HTML]{000000} 86.1 & {\cellcolor[HTML]{FCAE12}} \color[HTML]{000000} 87.0 & {\cellcolor[HTML]{F57D15}} \color[HTML]{F1F1F1} 80.9 & {\cellcolor[HTML]{7A1D6D}} \color[HTML]{F1F1F1} 59.1 \\
Intrinsic Circumstance & {\cellcolor[HTML]{FCA108}} \color[HTML]{000000} 85.4 & {\cellcolor[HTML]{FB9706}} \color[HTML]{000000} 84.2 & {\cellcolor[HTML]{FCAC11}} \color[HTML]{000000} 86.6 & {\cellcolor[HTML]{F78410}} \color[HTML]{F1F1F1} 81.7 & {\cellcolor[HTML]{B1325A}} \color[HTML]{F1F1F1} 67.1 & {\cellcolor[HTML]{DD513A}} \color[HTML]{F1F1F1} 74.4 & {\cellcolor[HTML]{982766}} \color[HTML]{F1F1F1} 63.4 & {\cellcolor[HTML]{DD513A}} \color[HTML]{F1F1F1} 74.4 & {\cellcolor[HTML]{420A68}} \color[HTML]{F1F1F1} 51.2 & {\cellcolor[HTML]{540F6D}} \color[HTML]{F1F1F1} 53.7 & {\cellcolor[HTML]{B93556}} \color[HTML]{F1F1F1} 68.3 & {\cellcolor[HTML]{982766}} \color[HTML]{F1F1F1} 63.4 \\
Extrinsic Circumstance & {\cellcolor[HTML]{FCA50A}} \color[HTML]{000000} 85.9 & {\cellcolor[HTML]{FCA50A}} \color[HTML]{000000} 85.9 & {\cellcolor[HTML]{FCA50A}} \color[HTML]{000000} 85.9 & {\cellcolor[HTML]{FCA50A}} \color[HTML]{000000} 85.9 & {\cellcolor[HTML]{F1731D}} \color[HTML]{F1F1F1} 79.5 & {\cellcolor[HTML]{ED6925}} \color[HTML]{F1F1F1} 78.2 & {\cellcolor[HTML]{E35933}} \color[HTML]{F1F1F1} 75.6 & {\cellcolor[HTML]{D74B3F}} \color[HTML]{F1F1F1} 73.1 & {\cellcolor[HTML]{F1731D}} \color[HTML]{F1F1F1} 79.5 & {\cellcolor[HTML]{DD513A}} \color[HTML]{F1F1F1} 74.4 & {\cellcolor[HTML]{B73557}} \color[HTML]{F1F1F1} 68.0 & {\cellcolor[HTML]{781C6D}} \color[HTML]{F1F1F1} 59.0 \\
Coreference Error & {\cellcolor[HTML]{FCAC11}} \color[HTML]{000000} 86.7 & {\cellcolor[HTML]{F3E35A}} \color[HTML]{000000} 92.9 & {\cellcolor[HTML]{FCAC11}} \color[HTML]{000000} 86.7 & {\cellcolor[HTML]{C73E4C}} \color[HTML]{F1F1F1} 70.4 & {\cellcolor[HTML]{F98B0B}} \color[HTML]{F1F1F1} 82.7 & {\cellcolor[HTML]{F98B0B}} \color[HTML]{F1F1F1} 82.7 & {\cellcolor[HTML]{B3325A}} \color[HTML]{F1F1F1} 67.4 & {\cellcolor[HTML]{1F0C48}} \color[HTML]{F1F1F1} 46.9 & {\cellcolor[HTML]{D34743}} \color[HTML]{F1F1F1} 72.5 & {\cellcolor[HTML]{FCAC11}} \color[HTML]{000000} 86.7 & {\cellcolor[HTML]{64156E}} \color[HTML]{F1F1F1} 56.1 & {\cellcolor[HTML]{A52C60}} \color[HTML]{F1F1F1} 65.3 \\
\midrule
% Intrinsic Error	&	92.3	&	88.7	&	87.4	&	84.7	&	77.3	&	75.8	&	70.9	&	70.3	&	57.1	&	51.5	&	64.1	&	57.7	\\
% Extrinsic Error	&	93.3	&	92.6	&	88.9	&	88.1	&	85.9	&	80.7	&	81.4	&	74.0	&	77.7	&	76.2	&	71.0	&	59.5	\\
Intrinsic Error & {\cellcolor[HTML]{F4DD4F}} \color[HTML]{000000} 92.3* & {\cellcolor[HTML]{FBBE23}} \color[HTML]{000000} 88.7 & {\cellcolor[HTML]{FCB216}} \color[HTML]{000000} 87.4 & {\cellcolor[HTML]{FB9B06}} \color[HTML]{000000} 84.7 & {\cellcolor[HTML]{EA632A}} \color[HTML]{F1F1F1} 77.3 & {\cellcolor[HTML]{E45A31}} \color[HTML]{F1F1F1} 75.8 & {\cellcolor[HTML]{CA404A}} \color[HTML]{F1F1F1} 70.9 & {\cellcolor[HTML]{C63D4D}} \color[HTML]{F1F1F1} 70.3 & {\cellcolor[HTML]{6C186E}} \color[HTML]{F1F1F1} 57.1 & {\cellcolor[HTML]{440A68}} \color[HTML]{F1F1F1} 51.5 & {\cellcolor[HTML]{9D2964}} \color[HTML]{F1F1F1} 64.1 & {\cellcolor[HTML]{6F196E}} \color[HTML]{F1F1F1} 57.7 \\
Extrinsic Error & {\cellcolor[HTML]{F3E55D}} \color[HTML]{000000} 93.3 & {\cellcolor[HTML]{F4DF53}} \color[HTML]{000000} 92.6 & {\cellcolor[HTML]{FAC026}} \color[HTML]{000000} 88.9 & {\cellcolor[HTML]{FBB81D}} \color[HTML]{000000} 88.1 & {\cellcolor[HTML]{FCA50A}} \color[HTML]{000000} 85.9 & {\cellcolor[HTML]{F57B17}} \color[HTML]{F1F1F1} 80.7 & {\cellcolor[HTML]{F78212}} \color[HTML]{F1F1F1} 81.4 & {\cellcolor[HTML]{DB503B}} \color[HTML]{F1F1F1} 74.0 & {\cellcolor[HTML]{EB6628}} \color[HTML]{F1F1F1} 77.7 & {\cellcolor[HTML]{E55C30}} \color[HTML]{F1F1F1} 76.2 & {\cellcolor[HTML]{CA404A}} \color[HTML]{F1F1F1} 71.0 & {\cellcolor[HTML]{7C1D6D}} \color[HTML]{F1F1F1} 59.5 \\
         \bottomrule
    \end{tabular}
    \subcaption{Task 1}
    \label{tab:sensitivity1}
    \end{subtable}

    \begin{subtable}[t]{\linewidth}
    \centering
    \begin{tabular}{llccccccccccc}
        \toprule
	&	BART	&	QAFaEv	&	CoCo	&	BERT	&	BLEURT	&	DAE	&	QuEv	&	\textsc{SummaC}	&	R-2	&	BLEU	&	$Q^2$	&	\textsc{FactCC}	\\
\midrule
% Overall	&	93.4	&	85.7	&	84.7	&	82.1	&	77.6	&	75.5	&	75.5	&	73.0	&	68.9	&	66.8	&	65.8	&	48.0	\\
Overall & {\cellcolor[HTML]{F2EA69}} \color[HTML]{000000} 93.4** & {\cellcolor[HTML]{FCA60C}} \color[HTML]{000000} 85.7 & {\cellcolor[HTML]{FC9F07}} \color[HTML]{000000} 84.7 & {\cellcolor[HTML]{F8890C}} \color[HTML]{F1F1F1} 82.1 & {\cellcolor[HTML]{EC6726}} \color[HTML]{F1F1F1} 77.6 & {\cellcolor[HTML]{E45A31}} \color[HTML]{F1F1F1} 75.5 & {\cellcolor[HTML]{E45A31}} \color[HTML]{F1F1F1} 75.5 & {\cellcolor[HTML]{D84C3E}} \color[HTML]{F1F1F1} 73.0 & {\cellcolor[HTML]{C03A51}} \color[HTML]{F1F1F1} 68.9 & {\cellcolor[HTML]{B3325A}} \color[HTML]{F1F1F1} 66.8 & {\cellcolor[HTML]{AB2F5E}} \color[HTML]{F1F1F1} 65.8 & {\cellcolor[HTML]{2B0B57}} \color[HTML]{F1F1F1} 48.0 \\
\midrule
% Intrinsic Predicate	&	82.4	&	88.2	&	88.2	&	82.4	&	82.4	&	82.4	&	82.4	&	64.7	&	70.6	&	64.7	&	76.5	&	64.7	\\
% Extrinsic Predicate	&	92.9	&	92.9	&	89.3	&	85.7	&	75.0	&	64.3	&	67.9	&	89.3	&	71.4	&	53.6	&	57.1	&	42.9	\\
% Intrinsic Entity	&	96.4	&	78.6	&	78.6	&	82.1	&	67.9	&	64.3	&	60.7	&	78.6	&	53.6	&	50.0	&	42.9	&	39.3	\\
% Extrinsic Entity	&	95.2	&	88.7	&	85.5	&	80.7	&	80.7	&	79.0	&	80.7	&	79.0	&	69.4	&	72.6	&	67.7	&	45.2	\\
% Intrinsic Circumstance	&	90.9	&	81.8	&	81.8	&	72.7	&	72.7	&	63.6	&	77.3	&	59.1	&	59.1	&	77.3	&	68.2	&	63.6	\\
% Extrinsic Circumstance	&	97.0	&	78.8	&	87.9	&	87.9	&	81.8	&	93.9	&	75.8	&	54.6	&	81.8	&	75.8	&	81.8	&	48.5	\\
Intrinsic Predicate & {\cellcolor[HTML]{F98C0A}} \color[HTML]{F1F1F1} 82.4 & {\cellcolor[HTML]{FBBC21}} \color[HTML]{000000} 88.2 & {\cellcolor[HTML]{FBBC21}} \color[HTML]{000000} 88.2 & {\cellcolor[HTML]{F98C0A}} \color[HTML]{F1F1F1} 82.4 & {\cellcolor[HTML]{F98C0A}} \color[HTML]{F1F1F1} 82.4 & {\cellcolor[HTML]{F98C0A}} \color[HTML]{F1F1F1} 82.4 & {\cellcolor[HTML]{F98C0A}} \color[HTML]{F1F1F1} 82.4 & {\cellcolor[HTML]{A32C61}} \color[HTML]{F1F1F1} 64.7 & {\cellcolor[HTML]{CA404A}} \color[HTML]{F1F1F1} 70.6 & {\cellcolor[HTML]{A32C61}} \color[HTML]{F1F1F1} 64.7 & {\cellcolor[HTML]{E9612B}} \color[HTML]{F1F1F1} 76.5 & {\cellcolor[HTML]{A32C61}} \color[HTML]{F1F1F1} 64.7 \\
Extrinsic Predicate & {\cellcolor[HTML]{F3E55D}} \color[HTML]{000000} 92.9 & {\cellcolor[HTML]{F3E55D}} \color[HTML]{000000} 92.9 & {\cellcolor[HTML]{FAC62D}} \color[HTML]{000000} 89.3 & {\cellcolor[HTML]{FCA60C}} \color[HTML]{000000} 85.7 & {\cellcolor[HTML]{E25734}} \color[HTML]{F1F1F1} 75.0 & {\cellcolor[HTML]{A02A63}} \color[HTML]{F1F1F1} 64.3 & {\cellcolor[HTML]{B93556}} \color[HTML]{F1F1F1} 67.9 & {\cellcolor[HTML]{FAC62D}} \color[HTML]{000000} 89.3 & {\cellcolor[HTML]{CF4446}} \color[HTML]{F1F1F1} 71.4 & {\cellcolor[HTML]{550F6D}} \color[HTML]{F1F1F1} 53.6 & {\cellcolor[HTML]{6D186E}} \color[HTML]{F1F1F1} 57.1 & {\cellcolor[HTML]{0A0722}} \color[HTML]{F1F1F1} 42.9 \\
Intrinsic Entity & {\cellcolor[HTML]{F9FC9D}} \color[HTML]{000000} 96.4* & {\cellcolor[HTML]{F06F20}} \color[HTML]{F1F1F1} 78.6 & {\cellcolor[HTML]{F06F20}} \color[HTML]{F1F1F1} 78.6 & {\cellcolor[HTML]{F8890C}} \color[HTML]{F1F1F1} 82.1 & {\cellcolor[HTML]{B93556}} \color[HTML]{F1F1F1} 67.9 & {\cellcolor[HTML]{A02A63}} \color[HTML]{F1F1F1} 64.3 & {\cellcolor[HTML]{87216B}} \color[HTML]{F1F1F1} 60.7 & {\cellcolor[HTML]{F06F20}} \color[HTML]{F1F1F1} 78.6 & {\cellcolor[HTML]{550F6D}} \color[HTML]{F1F1F1} 53.6 & {\cellcolor[HTML]{3B0964}} \color[HTML]{F1F1F1} 50.0 & {\cellcolor[HTML]{0A0722}} \color[HTML]{F1F1F1} 42.9 & {\cellcolor[HTML]{000004}} \color[HTML]{F1F1F1} 39.3 \\
Extrinsic Entity & {\cellcolor[HTML]{F3F68A}} \color[HTML]{000000} 95.2 & {\cellcolor[HTML]{FAC228}} \color[HTML]{000000} 88.7 & {\cellcolor[HTML]{FCA50A}} \color[HTML]{000000} 85.5 & {\cellcolor[HTML]{F67E14}} \color[HTML]{F1F1F1} 80.7 & {\cellcolor[HTML]{F67E14}} \color[HTML]{F1F1F1} 80.7 & {\cellcolor[HTML]{F1731D}} \color[HTML]{F1F1F1} 79.0 & {\cellcolor[HTML]{F67E14}} \color[HTML]{F1F1F1} 80.7 & {\cellcolor[HTML]{F1731D}} \color[HTML]{F1F1F1} 79.0 & {\cellcolor[HTML]{C33B4F}} \color[HTML]{F1F1F1} 69.4 & {\cellcolor[HTML]{D54A41}} \color[HTML]{F1F1F1} 72.6 & {\cellcolor[HTML]{B93556}} \color[HTML]{F1F1F1} 67.7 & {\cellcolor[HTML]{180C3C}} \color[HTML]{F1F1F1} 45.2 \\
Intrinsic Circumstance & {\cellcolor[HTML]{F7D340}} \color[HTML]{000000} 90.9 & {\cellcolor[HTML]{F8870E}} \color[HTML]{F1F1F1} 81.8 & {\cellcolor[HTML]{F8870E}} \color[HTML]{F1F1F1} 81.8 & {\cellcolor[HTML]{D74B3F}} \color[HTML]{F1F1F1} 72.7 & {\cellcolor[HTML]{D74B3F}} \color[HTML]{F1F1F1} 72.7 & {\cellcolor[HTML]{9B2964}} \color[HTML]{F1F1F1} 63.6 & {\cellcolor[HTML]{EB6628}} \color[HTML]{F1F1F1} 77.3 & {\cellcolor[HTML]{7C1D6D}} \color[HTML]{F1F1F1} 59.1 & {\cellcolor[HTML]{7C1D6D}} \color[HTML]{F1F1F1} 59.1 & {\cellcolor[HTML]{EB6628}} \color[HTML]{F1F1F1} 77.3 & {\cellcolor[HTML]{BC3754}} \color[HTML]{F1F1F1} 68.2 & {\cellcolor[HTML]{9B2964}} \color[HTML]{F1F1F1} 63.6 \\
Extrinsic Circumstance & {\cellcolor[HTML]{FCFFA4}} \color[HTML]{000000} 97.0 & {\cellcolor[HTML]{F1711F}} \color[HTML]{F1F1F1} 78.8 & {\cellcolor[HTML]{FBBA1F}} \color[HTML]{000000} 87.9 & {\cellcolor[HTML]{FBBA1F}} \color[HTML]{000000} 87.9 & {\cellcolor[HTML]{F8870E}} \color[HTML]{F1F1F1} 81.8 & {\cellcolor[HTML]{F1ED71}} \color[HTML]{000000} 93.9 & {\cellcolor[HTML]{E55C30}} \color[HTML]{F1F1F1} 75.8 & {\cellcolor[HTML]{5C126E}} \color[HTML]{F1F1F1} 54.6 & {\cellcolor[HTML]{F8870E}} \color[HTML]{F1F1F1} 81.8 & {\cellcolor[HTML]{E55C30}} \color[HTML]{F1F1F1} 75.8 & {\cellcolor[HTML]{F8870E}} \color[HTML]{F1F1F1} 81.8 & {\cellcolor[HTML]{2F0A5B}} \color[HTML]{F1F1F1} 48.5 \\
\midrule
% Intrinsic Error	&	91.0	&	82.1	&	82.1	&	79.1	&	73.1	&	68.7	&	71.6	&	68.7	&	59.7	&	62.7	&	59.7	&	53.7	\\
% Extrinsic Error	&	95.1	&	87.0	&	87.0	&	83.7	&	79.7	&	79.7	&	76.4	&	74.8	&	73.2	&	69.1	&	69.1	&	45.5	\\
Intrinsic Error & {\cellcolor[HTML]{F6D543}} \color[HTML]{000000} 91.0 & {\cellcolor[HTML]{F8890C}} \color[HTML]{F1F1F1} 82.1 & {\cellcolor[HTML]{F8890C}} \color[HTML]{F1F1F1} 82.1 & {\cellcolor[HTML]{F1731D}} \color[HTML]{F1F1F1} 79.1 & {\cellcolor[HTML]{D84C3E}} \color[HTML]{F1F1F1} 73.1 & {\cellcolor[HTML]{BF3952}} \color[HTML]{F1F1F1} 68.7 & {\cellcolor[HTML]{D04545}} \color[HTML]{F1F1F1} 71.6 & {\cellcolor[HTML]{BF3952}} \color[HTML]{F1F1F1} 68.7 & {\cellcolor[HTML]{801F6C}} \color[HTML]{F1F1F1} 59.7 & {\cellcolor[HTML]{952667}} \color[HTML]{F1F1F1} 62.7 & {\cellcolor[HTML]{801F6C}} \color[HTML]{F1F1F1} 59.7 & {\cellcolor[HTML]{550F6D}} \color[HTML]{F1F1F1} 53.7 \\
Extrinsic Error & {\cellcolor[HTML]{F3F586}} \color[HTML]{000000} 95.1** & {\cellcolor[HTML]{FCB216}} \color[HTML]{000000} 87.0 & {\cellcolor[HTML]{FCB216}} \color[HTML]{000000} 87.0 & {\cellcolor[HTML]{FB9606}} \color[HTML]{000000} 83.7 & {\cellcolor[HTML]{F37819}} \color[HTML]{F1F1F1} 79.7 & {\cellcolor[HTML]{F37819}} \color[HTML]{F1F1F1} 79.7 & {\cellcolor[HTML]{E8602D}} \color[HTML]{F1F1F1} 76.4 & {\cellcolor[HTML]{E15635}} \color[HTML]{F1F1F1} 74.8 & {\cellcolor[HTML]{D94D3D}} \color[HTML]{F1F1F1} 73.2 & {\cellcolor[HTML]{C13A50}} \color[HTML]{F1F1F1} 69.1 & {\cellcolor[HTML]{C13A50}} \color[HTML]{F1F1F1} 69.1 & {\cellcolor[HTML]{190C3E}} \color[HTML]{F1F1F1} 45.5 \\
%With numbers	&	95.4	&	83.7	&	84.5	&	86.1	&	77.5	&	76.7	&	77.5	&	70.5	&	69.8	&	68.2	&	65.9	&	49.6	\\
%Without numbers	&	89.6	&	89.6	&	85.1	&	74.6	&	77.6	&	73.1	&	71.6	&	77.6	&	67.2	&	64.2	&	65.7	&	44.8	\\
% With numbers & {\cellcolor[HTML]{F3F68A}} \color[HTML]{000000} 95.4 & {\cellcolor[HTML]{FB9606}} \color[HTML]{000000} 83.7 & {\cellcolor[HTML]{FB9D07}} \color[HTML]{000000} 84.5 & {\cellcolor[HTML]{FCAA0F}} \color[HTML]{000000} 86.1 & {\cellcolor[HTML]{EC6726}} \color[HTML]{F1F1F1} 77.5 & {\cellcolor[HTML]{E9612B}} \color[HTML]{F1F1F1} 76.7 & {\cellcolor[HTML]{EC6726}} \color[HTML]{F1F1F1} 77.5 & {\cellcolor[HTML]{CA404A}} \color[HTML]{F1F1F1} 70.5 & {\cellcolor[HTML]{C63D4D}} \color[HTML]{F1F1F1} 69.8 & {\cellcolor[HTML]{BC3754}} \color[HTML]{F1F1F1} 68.2 & {\cellcolor[HTML]{AD305D}} \color[HTML]{F1F1F1} 65.9 & {\cellcolor[HTML]{380962}} \color[HTML]{F1F1F1} 49.6 \\
% Without numbers & {\cellcolor[HTML]{F9C932}} \color[HTML]{000000} 89.6 & {\cellcolor[HTML]{F9C932}} \color[HTML]{000000} 89.6 & {\cellcolor[HTML]{FCA309}} \color[HTML]{000000} 85.1 & {\cellcolor[HTML]{E05536}} \color[HTML]{F1F1F1} 74.6 & {\cellcolor[HTML]{EC6726}} \color[HTML]{F1F1F1} 77.6 & {\cellcolor[HTML]{D84C3E}} \color[HTML]{F1F1F1} 73.1 & {\cellcolor[HTML]{D04545}} \color[HTML]{F1F1F1} 71.6 & {\cellcolor[HTML]{EC6726}} \color[HTML]{F1F1F1} 77.6 & {\cellcolor[HTML]{B43359}} \color[HTML]{F1F1F1} 67.2 & {\cellcolor[HTML]{A02A63}} \color[HTML]{F1F1F1} 64.2 & {\cellcolor[HTML]{AB2F5E}} \color[HTML]{F1F1F1} 65.7 & {\cellcolor[HTML]{150B37}} \color[HTML]{F1F1F1} 44.8 \\
         \bottomrule
    \end{tabular}  
    \subcaption{Task 2}
    \label{tab:sensitivity2}
    \end{subtable}
    \caption{
    {\bf Consistency ($\%$) of faithfulness evaluation metrics.} BART: \textsc{BARTScore}, QAFaEv: \textsc{QAFactEval}, BERT: \textsc{BERTScore}, QuEv: \textsc{QuestEval}, R-2: \textsc{ROUGE}-2. 
    %The highest and second highest score in each row are in bold and underscore, respectively, where s
    All values are color-coded. For each row,
    $*$ ($p < 0.05$) and $**$ ($p < 0.01$) indicate the results are statistically significant when comparing the best to the second-best metric. 
    }
    \label{tab:sensitivity}
\end{table*}

\paragraph{Consistency.}
%\aoife{I would avoid saying ``all metrics... except for...''. I think ``most metrics'' or ``all but XX metric'' are less misleading, especially when there's lots of text before the ``except''.}
The consistency studies of the two tasks\footnote{Note that for Task~2, the error types with only a few samples (e.g., Coreference and Other) are \emph{not} analyzed separately.} for all the metrics are reported in Table~\ref{tab:sensitivity}. 
In terms of the difficulty per error type, 1) for Task~1, \textit{Extrinsic Entity Error}s are generally the easiest,  while all but \textsc{BARTScore} struggle with \textit{Intrinsic Predicate Error}s; 2) for Task~2, \textit{Intrinsic Entity Error}s are the hardest. This implies that when annotators are not presented with any error types, the introduced error styles may differ from those in Task~1 (see Section \ref{sec:dataset_analysis}), potentially causing inconsistencies for metrics in these two tasks. Nevertheless, we observe that for both tasks, \textit{Intrinsic Error}s are more challenging than extrinsic ones across all but \textsc{FactCC} in Task~2. %This aligns with our intuition as extrinsic errors contain words that do not exist in the original article, while intrinsic ones are derivable from original articles, thus making them more subtle to be identified. 
This is likely because \textit{Intrinsic Error}s can be derived from the original article, while \textit{Extrinsic Error}s contain words that do not appear in the original article,  making \textit{Intrinsic Error}s more subtle to be identified than extrinsic ones.

For the overall performance (all error types are considered), \textsc{BARTScore} has the highest consistency in both tasks, though \textsc{BARTScore} has not been proposed specifically for faithfulness evaluations. Other metrics that rank top~4 in both tasks include \textsc{QAFactEval} and \textsc{CoCo}. By comparison, \textsc{$Q^2$} and \textsc{FactCC} have the worst consistency, even worse than $n$-gram-based metrics \textsc{ROUGE} and \textsc{BLEU}; nevertheless, they exhibit different rankings in terms of ROC AUC (see the next section). 

%\shuyang{I think it's better to use latex to produce the table?}\lm{absolutely. that is just a placeholder; it will be replaced once we figure out what to show and what goes to the appendix.}

\begin{table*}[t]
\centering
\scriptsize
\begin{subtable}[t]{\linewidth}
\centering
\begin{tabular}{llccccccccccc}
\toprule
 & QAFaEv & $Q^2$ & DAE & QuEv & BART & \textsc{FactCC} & CoCo & \textsc{SummaC} & BLEURT & BERT & R-2 & BLEU \\
\midrule
Overall & {\cellcolor[HTML]{FA9207}} \color[HTML]{000000} 71.5** & {\cellcolor[HTML]{BC3754}} \color[HTML]{F1F1F1} 64.2 & {\cellcolor[HTML]{B63458}} \color[HTML]{F1F1F1} 63.7 & {\cellcolor[HTML]{9F2A63}} \color[HTML]{F1F1F1} 62.0 & {\cellcolor[HTML]{84206B}} \color[HTML]{F1F1F1} 60.1 & {\cellcolor[HTML]{5A116E}} \color[HTML]{F1F1F1} 57.2 & {\cellcolor[HTML]{4F0D6C}} \color[HTML]{F1F1F1} 56.4 & {\cellcolor[HTML]{470B6A}} \color[HTML]{F1F1F1} 55.9 & {\cellcolor[HTML]{3B0964}} \color[HTML]{F1F1F1} 55.1 & {\cellcolor[HTML]{390963}} \color[HTML]{F1F1F1} 55.0 & {\cellcolor[HTML]{1E0C45}} \color[HTML]{F1F1F1} 53.2 & {\cellcolor[HTML]{030210}} \color[HTML]{F1F1F1} 50.6 \\
\midrule
Intrinsic Predicate & {\cellcolor[HTML]{DA4E3C}} \color[HTML]{F1F1F1} 66.7** & {\cellcolor[HTML]{57106E}} \color[HTML]{F1F1F1} 57.0 & {\cellcolor[HTML]{87216B}} \color[HTML]{F1F1F1} 60.4 & {\cellcolor[HTML]{490B6A}} \color[HTML]{F1F1F1} 56.0 & {\cellcolor[HTML]{8C2369}} \color[HTML]{F1F1F1} 60.7 & {\cellcolor[HTML]{010106}} \color[HTML]{F1F1F1} 50.1 & {\cellcolor[HTML]{390963}} \color[HTML]{F1F1F1} 55.0 & {\cellcolor[HTML]{210C4A}} \color[HTML]{F1F1F1} 53.4 & {\cellcolor[HTML]{2F0A5B}} \color[HTML]{F1F1F1} 54.3 & {\cellcolor[HTML]{3D0965}} \color[HTML]{F1F1F1} 55.2 & {\cellcolor[HTML]{0B0724}} \color[HTML]{F1F1F1} 51.6 & {\cellcolor[HTML]{02010A}} \color[HTML]{F1F1F1} 50.3 \\
Extrinsic Predicate & {\cellcolor[HTML]{FCB418}} \color[HTML]{000000} 73.5** & {\cellcolor[HTML]{7C1D6D}} \color[HTML]{F1F1F1} 59.6 & {\cellcolor[HTML]{6F196E}} \color[HTML]{F1F1F1} 58.7 & {\cellcolor[HTML]{7F1E6C}} \color[HTML]{F1F1F1} 59.8 & {\cellcolor[HTML]{67166E}} \color[HTML]{F1F1F1} 58.1 & {\cellcolor[HTML]{040312}} \color[HTML]{F1F1F1} 50.8 & {\cellcolor[HTML]{420A68}} \color[HTML]{F1F1F1} 55.5 & {\cellcolor[HTML]{5D126E}} \color[HTML]{F1F1F1} 57.4 & {\cellcolor[HTML]{2D0B59}} \color[HTML]{F1F1F1} 54.2 & {\cellcolor[HTML]{540F6D}} \color[HTML]{F1F1F1} 56.8 & {\cellcolor[HTML]{1F0C48}} \color[HTML]{F1F1F1} 53.3 & {\cellcolor[HTML]{02020E}} \color[HTML]{F1F1F1} 50.5 \\
Intrinsic Entity & {\cellcolor[HTML]{F4F88E}} \color[HTML]{000000} 77.7* & {\cellcolor[HTML]{FB9606}} \color[HTML]{000000} 71.7 & {\cellcolor[HTML]{F57B17}} \color[HTML]{F1F1F1} 70.1 & {\cellcolor[HTML]{E55C30}} \color[HTML]{F1F1F1} 67.8 & {\cellcolor[HTML]{B93556}} \color[HTML]{F1F1F1} 63.9 & {\cellcolor[HTML]{AB2F5E}} \color[HTML]{F1F1F1} 62.9 & {\cellcolor[HTML]{6A176E}} \color[HTML]{F1F1F1} 58.4 & {\cellcolor[HTML]{65156E}} \color[HTML]{F1F1F1} 58.0 & {\cellcolor[HTML]{490B6A}} \color[HTML]{F1F1F1} 56.0 & {\cellcolor[HTML]{4D0D6C}} \color[HTML]{F1F1F1} 56.3 & {\cellcolor[HTML]{1C0C43}} \color[HTML]{F1F1F1} 53.1 & {\cellcolor[HTML]{030210}} \color[HTML]{F1F1F1} 50.6 \\
Extrinsic Entity & {\cellcolor[HTML]{FCFFA4}} \color[HTML]{000000} 78.4* & {\cellcolor[HTML]{FCAC11}} \color[HTML]{000000} 73.1 & {\cellcolor[HTML]{E45A31}} \color[HTML]{F1F1F1} 67.7 & {\cellcolor[HTML]{F98C0A}} \color[HTML]{F1F1F1} 71.2 & {\cellcolor[HTML]{B3325A}} \color[HTML]{F1F1F1} 63.5 & {\cellcolor[HTML]{57106E}} \color[HTML]{F1F1F1} 57.0 & {\cellcolor[HTML]{6D186E}} \color[HTML]{F1F1F1} 58.6 & {\cellcolor[HTML]{5D126E}} \color[HTML]{F1F1F1} 57.5 & {\cellcolor[HTML]{6F196E}} \color[HTML]{F1F1F1} 58.7 & {\cellcolor[HTML]{540F6D}} \color[HTML]{F1F1F1} 56.8 & {\cellcolor[HTML]{400A67}} \color[HTML]{F1F1F1} 55.4 & {\cellcolor[HTML]{040312}} \color[HTML]{F1F1F1} 50.8 \\
Intrinsic Circumstance & {\cellcolor[HTML]{F98E09}} \color[HTML]{F1F1F1} 71.3* & {\cellcolor[HTML]{AB2F5E}} \color[HTML]{F1F1F1} 62.9 & {\cellcolor[HTML]{B63458}} \color[HTML]{F1F1F1} 63.7 & {\cellcolor[HTML]{3E0966}} \color[HTML]{F1F1F1} 55.3 & {\cellcolor[HTML]{400A67}} \color[HTML]{F1F1F1} 55.4 & {\cellcolor[HTML]{9D2964}} \color[HTML]{F1F1F1} 61.9 & {\cellcolor[HTML]{260C51}} \color[HTML]{F1F1F1} 53.8 & {\cellcolor[HTML]{5A116E}} \color[HTML]{F1F1F1} 57.2 & {\cellcolor[HTML]{110A30}} \color[HTML]{F1F1F1} 52.2 & {\cellcolor[HTML]{050417}} \color[HTML]{F1F1F1} 51.0 & {\cellcolor[HTML]{0D0829}} \color[HTML]{F1F1F1} 51.9 & {\cellcolor[HTML]{000004}} \color[HTML]{F1F1F1} 49.8 \\
Extrinsic Circumstance & {\cellcolor[HTML]{FBBA1F}} \color[HTML]{000000} 73.9** & {\cellcolor[HTML]{BA3655}} \color[HTML]{F1F1F1} 64.0 & {\cellcolor[HTML]{902568}} \color[HTML]{F1F1F1} 61.0 & {\cellcolor[HTML]{5D126E}} \color[HTML]{F1F1F1} 57.5 & {\cellcolor[HTML]{57106E}} \color[HTML]{F1F1F1} 57.0 & {\cellcolor[HTML]{71196E}} \color[HTML]{F1F1F1} 58.8 & {\cellcolor[HTML]{360961}} \color[HTML]{F1F1F1} 54.8 & {\cellcolor[HTML]{781C6D}} \color[HTML]{F1F1F1} 59.3 & {\cellcolor[HTML]{210C4A}} \color[HTML]{F1F1F1} 53.4 & {\cellcolor[HTML]{140B34}} \color[HTML]{F1F1F1} 52.4 & {\cellcolor[HTML]{3B0964}} \color[HTML]{F1F1F1} 55.1 & {\cellcolor[HTML]{030210}} \color[HTML]{F1F1F1} 50.6 \\
Coreference Error & {\cellcolor[HTML]{69166E}} \color[HTML]{F1F1F1} 58.2 & {\cellcolor[HTML]{57106E}} \color[HTML]{F1F1F1} 57.0 & {\cellcolor[HTML]{8D2369}} \color[HTML]{F1F1F1} 60.8 & {\cellcolor[HTML]{A02A63}} \color[HTML]{F1F1F1} 62.1 & {\cellcolor[HTML]{7D1E6D}} \color[HTML]{F1F1F1} 59.7 & {\cellcolor[HTML]{65156E}} \color[HTML]{F1F1F1} 58.0 & {\cellcolor[HTML]{5A116E}} \color[HTML]{F1F1F1} 57.2 & {\cellcolor[HTML]{010106}} \color[HTML]{F1F1F1} 50.1 & {\cellcolor[HTML]{400A67}} \color[HTML]{F1F1F1} 55.4 & {\cellcolor[HTML]{3E0966}} \color[HTML]{F1F1F1} 55.3 & {\cellcolor[HTML]{1B0C41}} \color[HTML]{F1F1F1} 53.0 & {\cellcolor[HTML]{0C0826}} \color[HTML]{F1F1F1} 51.7 \\
\midrule
Intrinsic Error & {\cellcolor[HTML]{FB9B06}} \color[HTML]{000000} 72.1** & {\cellcolor[HTML]{BF3952}} \color[HTML]{F1F1F1} 64.4 & {\cellcolor[HTML]{C63D4D}} \color[HTML]{F1F1F1} 64.9 & {\cellcolor[HTML]{87216B}} \color[HTML]{F1F1F1} 60.4 & {\cellcolor[HTML]{88226A}} \color[HTML]{F1F1F1} 60.5 & {\cellcolor[HTML]{67166E}} \color[HTML]{F1F1F1} 58.1 & {\cellcolor[HTML]{490B6A}} \color[HTML]{F1F1F1} 56.0 & {\cellcolor[HTML]{4A0C6B}} \color[HTML]{F1F1F1} 56.1 & {\cellcolor[HTML]{310A5C}} \color[HTML]{F1F1F1} 54.4 & {\cellcolor[HTML]{320A5E}} \color[HTML]{F1F1F1} 54.6 & {\cellcolor[HTML]{120A32}} \color[HTML]{F1F1F1} 52.3 & {\cellcolor[HTML]{02010A}} \color[HTML]{F1F1F1} 50.3 \\
Extrinsic Error & {\cellcolor[HTML]{F6D746}} \color[HTML]{000000} 75.6** & {\cellcolor[HTML]{D84C3E}} \color[HTML]{F1F1F1} 66.5 & {\cellcolor[HTML]{AE305C}} \color[HTML]{F1F1F1} 63.2 & {\cellcolor[HTML]{B93556}} \color[HTML]{F1F1F1} 63.9 & {\cellcolor[HTML]{7F1E6C}} \color[HTML]{F1F1F1} 59.8 & {\cellcolor[HTML]{440A68}} \color[HTML]{F1F1F1} 55.7 & {\cellcolor[HTML]{4F0D6C}} \color[HTML]{F1F1F1} 56.5 & {\cellcolor[HTML]{64156E}} \color[HTML]{F1F1F1} 57.9 & {\cellcolor[HTML]{450A69}} \color[HTML]{F1F1F1} 55.8 & {\cellcolor[HTML]{400A67}} \color[HTML]{F1F1F1} 55.4 & {\cellcolor[HTML]{320A5E}} \color[HTML]{F1F1F1} 54.6 & {\cellcolor[HTML]{030210}} \color[HTML]{F1F1F1} 50.6 \\
\bottomrule
\end{tabular}
\subcaption{Task 1}
\label{tab:task1_auc}
\end{subtable}

\begin{subtable}[t]{\linewidth}
\centering
\begin{tabular}{llccccccccccc}
\toprule
 & QAFaEv & $Q^2$ & DAE & QuEv & BART & \textsc{SummaC} & CoCo & BERT & R-2 & BLEURT & \textsc{FactCC} & BLEU \\
% Error Type &  &  &  &  &  &  &  &  &  &  &  &  \\
\midrule
Overall & {\cellcolor[HTML]{F9C72F}} \color[HTML]{000000} 71.2** & {\cellcolor[HTML]{B93556}} \color[HTML]{F1F1F1} 61.3 & {\cellcolor[HTML]{922568}} \color[HTML]{F1F1F1} 58.8 & {\cellcolor[HTML]{7D1E6D}} \color[HTML]{F1F1F1} 57.4 & {\cellcolor[HTML]{7D1E6D}} \color[HTML]{F1F1F1} 57.4 & {\cellcolor[HTML]{751B6E}} \color[HTML]{F1F1F1} 56.9 & {\cellcolor[HTML]{510E6C}} \color[HTML]{F1F1F1} 54.5 & {\cellcolor[HTML]{4A0C6B}} \color[HTML]{F1F1F1} 54.1 & {\cellcolor[HTML]{490B6A}} \color[HTML]{F1F1F1} 54.0 & {\cellcolor[HTML]{310A5C}} \color[HTML]{F1F1F1} 52.6 & {\cellcolor[HTML]{1F0C48}} \color[HTML]{F1F1F1} 51.5 & {\cellcolor[HTML]{0E092B}} \color[HTML]{F1F1F1} 50.3 \\
\midrule
Intrinsic Predicate & {\cellcolor[HTML]{F5DB4C}} \color[HTML]{000000} 72.3 & {\cellcolor[HTML]{D34743}} \color[HTML]{F1F1F1} 63.3 & {\cellcolor[HTML]{AE305C}} \color[HTML]{F1F1F1} 60.6 & {\cellcolor[HTML]{7D1E6D}} \color[HTML]{F1F1F1} 57.4 & {\cellcolor[HTML]{57106E}} \color[HTML]{F1F1F1} 55.0 & {\cellcolor[HTML]{65156E}} \color[HTML]{F1F1F1} 55.9 & {\cellcolor[HTML]{62146E}} \color[HTML]{F1F1F1} 55.7 & {\cellcolor[HTML]{7D1E6D}} \color[HTML]{F1F1F1} 57.4 & {\cellcolor[HTML]{420A68}} \color[HTML]{F1F1F1} 53.6 & {\cellcolor[HTML]{540F6D}} \color[HTML]{F1F1F1} 54.7 & {\cellcolor[HTML]{922568}} \color[HTML]{F1F1F1} 58.8 & {\cellcolor[HTML]{210C4A}} \color[HTML]{F1F1F1} 51.6 \\
Extrinsic Predicate & {\cellcolor[HTML]{FCFFA4}} \color[HTML]{000000} 74.7** & {\cellcolor[HTML]{781C6D}} \color[HTML]{F1F1F1} 57.1 & {\cellcolor[HTML]{7F1E6C}} \color[HTML]{F1F1F1} 57.5 & {\cellcolor[HTML]{61136E}} \color[HTML]{F1F1F1} 55.6 & {\cellcolor[HTML]{A82E5F}} \color[HTML]{F1F1F1} 60.2 & {\cellcolor[HTML]{A22B62}} \color[HTML]{F1F1F1} 59.8 & {\cellcolor[HTML]{65156E}} \color[HTML]{F1F1F1} 55.9 & {\cellcolor[HTML]{67166E}} \color[HTML]{F1F1F1} 56.0 & {\cellcolor[HTML]{5A116E}} \color[HTML]{F1F1F1} 55.2 & {\cellcolor[HTML]{4D0D6C}} \color[HTML]{F1F1F1} 54.3 & {\cellcolor[HTML]{000004}} \color[HTML]{F1F1F1} 48.3 & {\cellcolor[HTML]{0E092B}} \color[HTML]{F1F1F1} 50.3 \\
Intrinsic Entity & {\cellcolor[HTML]{E8602D}} \color[HTML]{F1F1F1} 65.3 & {\cellcolor[HTML]{71196E}} \color[HTML]{F1F1F1} 56.6 & {\cellcolor[HTML]{84206B}} \color[HTML]{F1F1F1} 57.8 & {\cellcolor[HTML]{5F136E}} \color[HTML]{F1F1F1} 55.5 & {\cellcolor[HTML]{A82E5F}} \color[HTML]{F1F1F1} 60.2 & {\cellcolor[HTML]{82206C}} \color[HTML]{F1F1F1} 57.7 & {\cellcolor[HTML]{82206C}} \color[HTML]{F1F1F1} 57.7 & {\cellcolor[HTML]{69166E}} \color[HTML]{F1F1F1} 56.1 & {\cellcolor[HTML]{2F0A5B}} \color[HTML]{F1F1F1} 52.5 & {\cellcolor[HTML]{310A5C}} \color[HTML]{F1F1F1} 52.6 & {\cellcolor[HTML]{0E092B}} \color[HTML]{F1F1F1} 50.3 & {\cellcolor[HTML]{10092D}} \color[HTML]{F1F1F1} 50.4 \\
Extrinsic Entity & {\cellcolor[HTML]{FAFDA1}} \color[HTML]{000000} 74.5** & {\cellcolor[HTML]{E25734}} \color[HTML]{F1F1F1} 64.6 & {\cellcolor[HTML]{88226A}} \color[HTML]{F1F1F1} 58.1 & {\cellcolor[HTML]{AB2F5E}} \color[HTML]{F1F1F1} 60.4 & {\cellcolor[HTML]{902568}} \color[HTML]{F1F1F1} 58.7 & {\cellcolor[HTML]{9F2A63}} \color[HTML]{F1F1F1} 59.6 & {\cellcolor[HTML]{5C126E}} \color[HTML]{F1F1F1} 55.3 & {\cellcolor[HTML]{4A0C6B}} \color[HTML]{F1F1F1} 54.1 & {\cellcolor[HTML]{550F6D}} \color[HTML]{F1F1F1} 54.8 & {\cellcolor[HTML]{380962}} \color[HTML]{F1F1F1} 53.0 & {\cellcolor[HTML]{08051D}} \color[HTML]{F1F1F1} 49.7 & {\cellcolor[HTML]{140B34}} \color[HTML]{F1F1F1} 50.7 \\
Intrinsic Circumstance & {\cellcolor[HTML]{E75E2E}} \color[HTML]{F1F1F1} 65.2 & {\cellcolor[HTML]{771C6D}} \color[HTML]{F1F1F1} 57.0 & {\cellcolor[HTML]{5D126E}} \color[HTML]{F1F1F1} 55.4 & {\cellcolor[HTML]{64156E}} \color[HTML]{F1F1F1} 55.8 & {\cellcolor[HTML]{71196E}} \color[HTML]{F1F1F1} 56.6 & {\cellcolor[HTML]{65156E}} \color[HTML]{F1F1F1} 55.9 & {\cellcolor[HTML]{470B6A}} \color[HTML]{F1F1F1} 53.9 & {\cellcolor[HTML]{240C4F}} \color[HTML]{F1F1F1} 51.9 & {\cellcolor[HTML]{2D0B59}} \color[HTML]{F1F1F1} 52.4 & {\cellcolor[HTML]{240C4F}} \color[HTML]{F1F1F1} 51.9 & {\cellcolor[HTML]{1E0C45}} \color[HTML]{F1F1F1} 51.4 & {\cellcolor[HTML]{1C0C43}} \color[HTML]{F1F1F1} 51.3 \\
Extrinsic Circumstance & {\cellcolor[HTML]{F7D340}} \color[HTML]{000000} 71.9 & {\cellcolor[HTML]{EE6A24}} \color[HTML]{F1F1F1} 66.0 & {\cellcolor[HTML]{F98B0B}} \color[HTML]{F1F1F1} 67.9 & {\cellcolor[HTML]{7C1D6D}} \color[HTML]{F1F1F1} 57.3 & {\cellcolor[HTML]{85216B}} \color[HTML]{F1F1F1} 57.9 & {\cellcolor[HTML]{4A0C6B}} \color[HTML]{F1F1F1} 54.1 & {\cellcolor[HTML]{57106E}} \color[HTML]{F1F1F1} 55.0 & {\cellcolor[HTML]{450A69}} \color[HTML]{F1F1F1} 53.8 & {\cellcolor[HTML]{5D126E}} \color[HTML]{F1F1F1} 55.4 & {\cellcolor[HTML]{510E6C}} \color[HTML]{F1F1F1} 54.5 & {\cellcolor[HTML]{61136E}} \color[HTML]{F1F1F1} 55.6 & {\cellcolor[HTML]{1B0C41}} \color[HTML]{F1F1F1} 51.2 \\
\midrule
Intrinsic Error & {\cellcolor[HTML]{F1731D}} \color[HTML]{F1F1F1} 66.5** & {\cellcolor[HTML]{801F6C}} \color[HTML]{F1F1F1} 57.6 & {\cellcolor[HTML]{7F1E6C}} \color[HTML]{F1F1F1} 57.5 & {\cellcolor[HTML]{5D126E}} \color[HTML]{F1F1F1} 55.4 & {\cellcolor[HTML]{62146E}} \color[HTML]{F1F1F1} 55.7 & {\cellcolor[HTML]{67166E}} \color[HTML]{F1F1F1} 56.0 & {\cellcolor[HTML]{520E6D}} \color[HTML]{F1F1F1} 54.6 & {\cellcolor[HTML]{450A69}} \color[HTML]{F1F1F1} 53.8 & {\cellcolor[HTML]{290B55}} \color[HTML]{F1F1F1} 52.2 & {\cellcolor[HTML]{1F0C48}} \color[HTML]{F1F1F1} 51.5 & {\cellcolor[HTML]{320A5E}} \color[HTML]{F1F1F1} 52.7 & {\cellcolor[HTML]{0E092B}} \color[HTML]{F1F1F1} 50.3 \\
Extrinsic Error & {\cellcolor[HTML]{F1F179}} \color[HTML]{000000} 73.5** & {\cellcolor[HTML]{D44842}} \color[HTML]{F1F1F1} 63.4 & {\cellcolor[HTML]{A22B62}} \color[HTML]{F1F1F1} 59.8 & {\cellcolor[HTML]{8C2369}} \color[HTML]{F1F1F1} 58.4 & {\cellcolor[HTML]{8D2369}} \color[HTML]{F1F1F1} 58.5 & {\cellcolor[HTML]{85216B}} \color[HTML]{F1F1F1} 57.9 & {\cellcolor[HTML]{550F6D}} \color[HTML]{F1F1F1} 54.8 & {\cellcolor[HTML]{4A0C6B}} \color[HTML]{F1F1F1} 54.1 & {\cellcolor[HTML]{540F6D}} \color[HTML]{F1F1F1} 54.7 & {\cellcolor[HTML]{3E0966}} \color[HTML]{F1F1F1} 53.4 & {\cellcolor[HTML]{160B39}} \color[HTML]{F1F1F1} 50.9 & {\cellcolor[HTML]{10092D}} \color[HTML]{F1F1F1} 50.4 \\
%\midrule
%With numbers & {\cellcolor[HTML]{FCB014}} \color[HTML]{000000} 70.0 & {\cellcolor[HTML]{B1325A}} \color[HTML]{F1F1F1} 60.8 & {\cellcolor[HTML]{8D2369}} \color[HTML]{F1F1F1} 58.5 & {\cellcolor[HTML]{6F196E}} \color[HTML]{F1F1F1} 56.5 & {\cellcolor[HTML]{7F1E6C}} \color[HTML]{F1F1F1} 57.5 & {\cellcolor[HTML]{71196E}} \color[HTML]{F1F1F1} 56.6 & {\cellcolor[HTML]{520E6D}} \color[HTML]{F1F1F1} 54.6 & {\cellcolor[HTML]{4A0C6B}} \color[HTML]{F1F1F1} 54.1 & {\cellcolor[HTML]{4F0D6C}} \color[HTML]{F1F1F1} 54.4 & {\cellcolor[HTML]{2B0B57}} \color[HTML]{F1F1F1} 52.3 & {\cellcolor[HTML]{290B55}} \color[HTML]{F1F1F1} 52.2 & {\cellcolor[HTML]{10092D}} \color[HTML]{F1F1F1} 50.4 \\
%Without numbers & {\cellcolor[HTML]{F2F482}} \color[HTML]{000000} 73.7 & {\cellcolor[HTML]{C63D4D}} \color[HTML]{F1F1F1} 62.3 & {\cellcolor[HTML]{972766}} \color[HTML]{F1F1F1} 59.1 & {\cellcolor[HTML]{952667}} \color[HTML]{F1F1F1} 59.0 & {\cellcolor[HTML]{8C2369}} \color[HTML]{F1F1F1} 58.4 & {\cellcolor[HTML]{88226A}} \color[HTML]{F1F1F1} 58.2 & {\cellcolor[HTML]{5C126E}} \color[HTML]{F1F1F1} 55.3 & {\cellcolor[HTML]{520E6D}} \color[HTML]{F1F1F1} 54.6 & {\cellcolor[HTML]{520E6D}} \color[HTML]{F1F1F1} 54.6 & {\cellcolor[HTML]{400A67}} \color[HTML]{F1F1F1} 53.5 & {\cellcolor[HTML]{110A30}} \color[HTML]{F1F1F1} 50.5 & {\cellcolor[HTML]{110A30}} \color[HTML]{F1F1F1} 50.5 \\
\bottomrule
\end{tabular}
\subcaption{Task 2}
\label{tab:task2_auc}    
\end{subtable}
\caption{
    {\bf ROC AUC (\%) of faithfulness evaluation metrics.} 
    BART: \textsc{BARTScore}, QAFaEv: \textsc{QAFactEval}, BERT: \textsc{BERTScore}, QuEv: \textsc{QuestEval}, R-2: \textsc{ROUGE}-2. 
    All values are color-coded. For  each row,
    $*$ ($p < 0.05$) and $**$ ($p < 0.01$) indicate the results are statistically significant when comparing the best to the second-best metric.}
\label{tab:auc}
\end{table*}

\paragraph{ROC AUC.}
% The ROC AUC study of detecting unfaithful summaries for all metrics are presented in 
ROC AUC scores are presented in Table~\ref{tab:auc}.
% \rob{From a narrative standpoint, it makes more sense to me to lead with how ROC vs. consistency results are different, as opposed to how they are similar. Right now we are starting with what I feel like are the least interesting results, and a claim that confusingly contradicts the narrative in the introduction (i.e., that consistency != discriminability).}
% First, when all error types are considered, 
We observe that the overall ranking of faithfulness metrics according to ROC AUC substantially differs from the ranking according to consistency.
In particular, the rank of \textsc{BARTScore} drops from the top one to the fifth, while \textsc{$Q^2$} improves significantly from second to last to  second overall.
\textsc{QAFactEval} consistently exhibits high performance and even ranks first under ROC AUC, while $n$-gram based metrics, e.g., \textsc{ROUGE}-2 and \textsc{BLEU} consistently show the worst performance, as expected.
In general, metrics that are specifically proposed for faithfulness evaluations rank higher than generic NLG evaluation metrics.
% Interestingly, th

We additionally observe that the relative rankings of ROC AUC scores across error types and task settings are largely consistent with the relative rankings of consistency scores.
Specifically, we again observe that on a per metric basis: 1) ROC AUC scores are generally lower for Task~2 than Task~1 (particularly for \textit{Entity Error}s), and 
%there is no obvious difference among error types for Task~2;
2) metrics generally show worse performance on \textit{Intrinsic Error}s than extrinsic ones.
%Interestingly, we notice that the ranking of metrics in terms of their overall ROC score substantially differs from their ranking according to consistency.

% and unfaithful summaries in Task~2 is more challenging than those in Task~1.
%(iv) metrics also find harder time to detect unfaithful summaries due to numbers. \lm{we do not have this obervation in consistency}
% By contrast, we observe that the identification difficulties between correct and incorrect annotator responses become less distinguishable (i.e., mostly around $1$--$2\%$), suggesting metrics' unfaithfulness identification capabiblity is insensitive to correct/incorrect annotator responses. Similar to the sensivitiy study, we report results on ROC AUC under our corrected error types and have similar observations; see Appendix~\ref{sec:appendix_additional_res} for details.

% Regarding the overall performance under ROC AUC, we observe that the ranking of metrics differs from  consistency. 
%As we can see, our used two meta-evaluation protocols provide different ranks of metrics. 
%Such different rankings under the two meta-evaluation protocols can be attributed to the nature of the two protocols. 
For our two meta-evaluation protocols, consistency is suitable for the pairwise ranking of two summaries for a given input article, while ROC AUC is more adequate in evaluating the absolute capacity of unfaithful summary detection. 
% Therefore, if a metric experiences high consistency while low ROC AUC, it implies that the predicted faithful and unfaithful scores on average are too close. Such score closeness results in overall score overlapping between the faithful and unfaithful summary groups, thus making it  difficult to
If a metric has high consistency but low ROC AUC, it implies that the scores for predicted faithful and unfaithful summaries overlap frequently. Such overlap makes it challenging to establish a clear decision boundary for classifications.
%Therefore, given a pairwise faithful and unfaithful summaries, a metric that experiences better pairwise ranking does not necessarily imply that it is able to 
% find a good decision boundary to separate these two groups. 
Hence, to improve the classification capability of metrics with high consistency, more calibration is needed to increase the score gap between faithful and unfaithful summaries.   

\section{Analysis of \dataset}
\label{sec:dataset_analysis}
In this section, we conduct more analysis of \dataset by studying how \dataset differs from other benchmarks, followed by a qualitative analysis of the detection difficulty between Tasks~1 and 2.

% \rob{
% The following paragraphs feel a bit out of place.
% The idea of taking probability differences was not brought up earlier in the section, and is more of a characterization of the benchmark than an 'evaluation of faithfulness evaluation metrics'.
% A similar claim could be made for the qualitative examples, especially if they are intended to illustrate how we corrected error type annotations, or they are illustrating annotation quality differences between task 1 and task 2.
% Accordingly, I think it could be sensible to rework the remained of this section into a seperate section that about "Analyzing \dataset" or "How \dataset differs from other benchmarks".
% }

\paragraph{Comparison with Model-Generated Unfaithful Summaries.}

\begin{table*}[tb]
    \scriptsize
    \centering
    \begin{tabular}{m{0.38\textwidth}m{0.145\textwidth}m{0.06\textwidth}m{0.145\textwidth}m{0.145\textwidth}}
        \toprule
        
        \multirow{2}{0.38\textwidth}{\centering Article (Partial)} & \multirow{2}{0.145\textwidth}{\centering Reference Summary} & \multirow{2}{0.06\textwidth}{\centering Error Type} & \multicolumn{2}{c}{Edited Summary} \tabularnewline
        % \cline{4-5}
        & & & \centering Task 1 & \centering Task 2 \tabularnewline
 % & Analysis\\
 
\midrule
... Detective Chief Inspector Paul Johnson of the London Metropolitan Police Flying Squad said the thieves appeared to have gained access to the vault of Hatton Garden Safe Deposit Ltd through the shaft of \textbf{an elevator that is used by several businesses in the building}. ...& Police say the thieves gained entry through the building's \textbf{communal} elevator shaft. ...& Extrinsic Entity Error & Police say the thieves gained entry through the building's communal \textbf{staircase}. ...& Police say the thieves gained entry through the building's \textbf{private} elevator shaft. ... \\
% & Great hallucination edits  \\
\midrule
... \textbf{Claire Nugent}, 43, and Nigel Morter, 47, have been married for 14 years... She said: `Every night I come home to my Sixties bubble, switch on my old \textbf{record player}, listen to some vinyl, and all the stresses of 2015 melt away' ...& \textbf{Claire Nugent} and Nigel Morter restored a ... likes to come home and switch on an \textbf{old record player} like in the 60s.&  Extrinsic Entity Error & \textbf{Tim Horton} and Nigel Morter restored a ... likes to come home and switch on an old record player like in the 60s.& Claire Nugent and Nigel Morter restored a ... likes to come home and switch on a \textbf{black and white TV} like in the 60s. \\
\midrule
Almost three years after nearly leaving Liverpool Jordan Henderson \textbf{has committed} his long-term future to the club convinced he can win silverware at Anfield. ... Henderson has \textbf{urged} Liverpool team-mate Raheem Sterling to follow his lead by signing a new deal.&Jordan Henderson \textbf{has signed} a new five-year deal at Anfield. ... Henderson has \textbf{urged} Raheem Sterling to ...&  Extrinsic Predicate Error &Jordan Henderson has signed a new five-year deal at Anfield. ... Henderson has \textbf{discouraged} Raheem Sterling to ...& Jordan Henderson \textbf{is considering signing} a new five-year deal at Anfield. ... Henderson has urged Raheem Sterling to ... \\

% & High faithful score from the metrics \\
         \bottomrule
    \end{tabular}  
    \caption{
        {\bf Qualitative examples illustrating the higher difficulty of edits in Task~2.}
        Each row contains a pair of edits from Tasks~1 and 2 pertaining to the same article and error type, where more metrics are inconsistent for  Task~2 edits.
    }
    \label{tab:casestudy}
\end{table*}
\begin{figure}[tb]
    \centering   
    \includegraphics[width=0.44\textwidth]{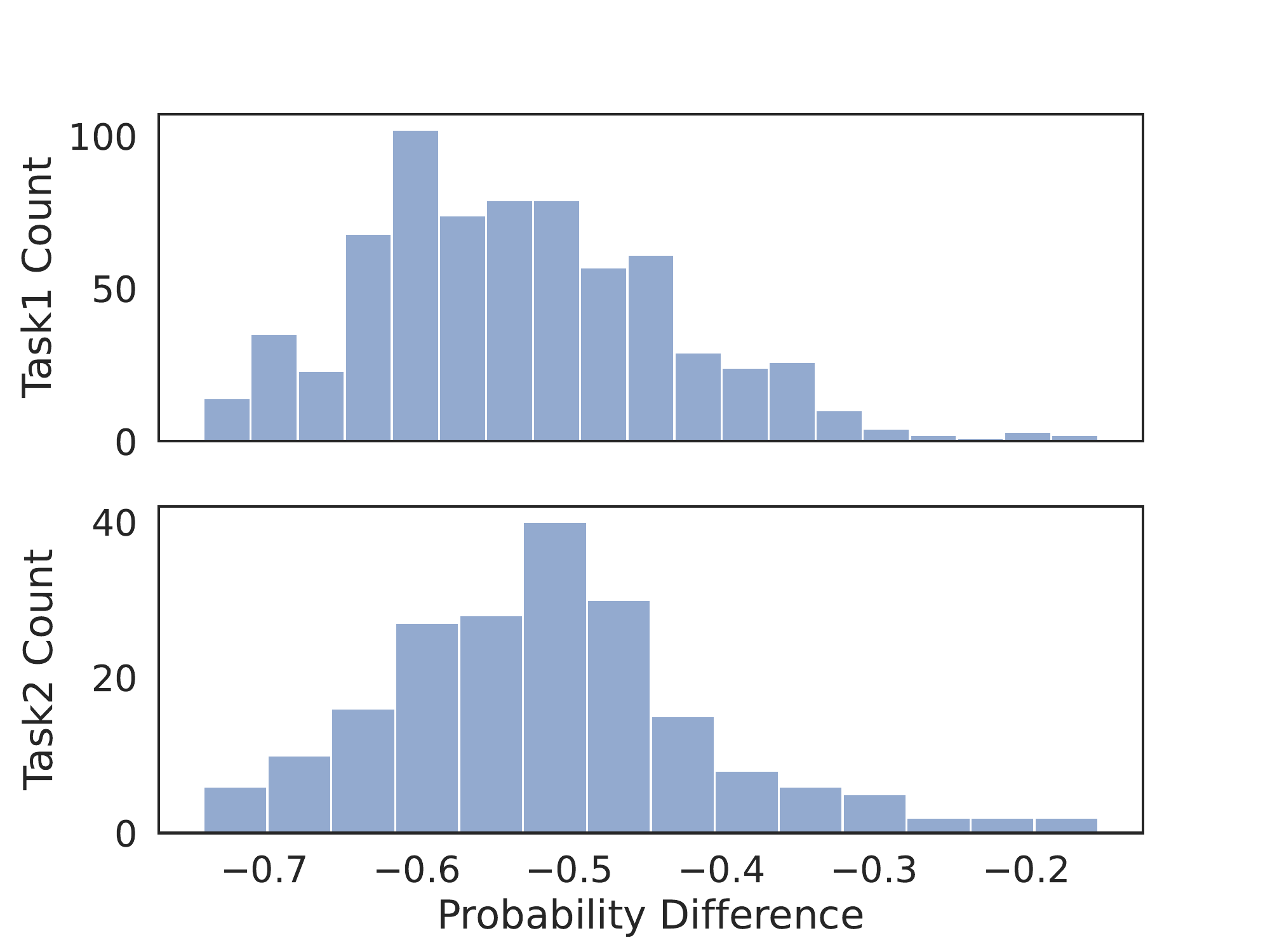}
    \caption{
        {\bf Distribution of probability differences between human-edited and model-generated summaries on Tasks~1 and 2}.
        Probabilities are computed using a BART-based summarizer.
        The high frequency of negative values indicates that human-edited summaries tend to have lower probabilities under the model.
    }
    \label{fig:prob_diff}
\end{figure}

%\rob{There's a bit of detail lacking in this paragraph - is same model that generated the summaries used to evaluation the probability? Also, is there a better experiment we could perform? E.g., measuring the probability of summaries in FRANK, TRUE, SummaC, etc. under this model?}\lm{there is only one model here. The one trained by the CNN/DM training data. Then we use it to check the probability for any given summary of an article. Is that right, Shuyang? Those datasets in FRANK are model generated, and it is likely the models are just trained by the CNN/DM training data?}
We compare the generation probabilities of our edited summaries to those of summaries generated from beam search by a BART-based summarizer (trained using the training data of CNN/DailyMail) for the same set of documents in our dataset.
We report the difference of these generation probabilities normalized by the text length in Figure~\ref{fig:prob_diff},
where we find our edited summaries are much different from model generations in terms of the model generation probabilities.
This highlights that existing metrics may not work well on summaries of various styles and experiments are needed to verify their effectiveness in human-generated unfaithful summaries.
%when applying to new datasets, \aoife{This last claus does not follow well from the previous text} and further confirms the contribution of \dataset.

\begin{figure}[tb]
    \centering   
\includegraphics[width=0.5\textwidth]{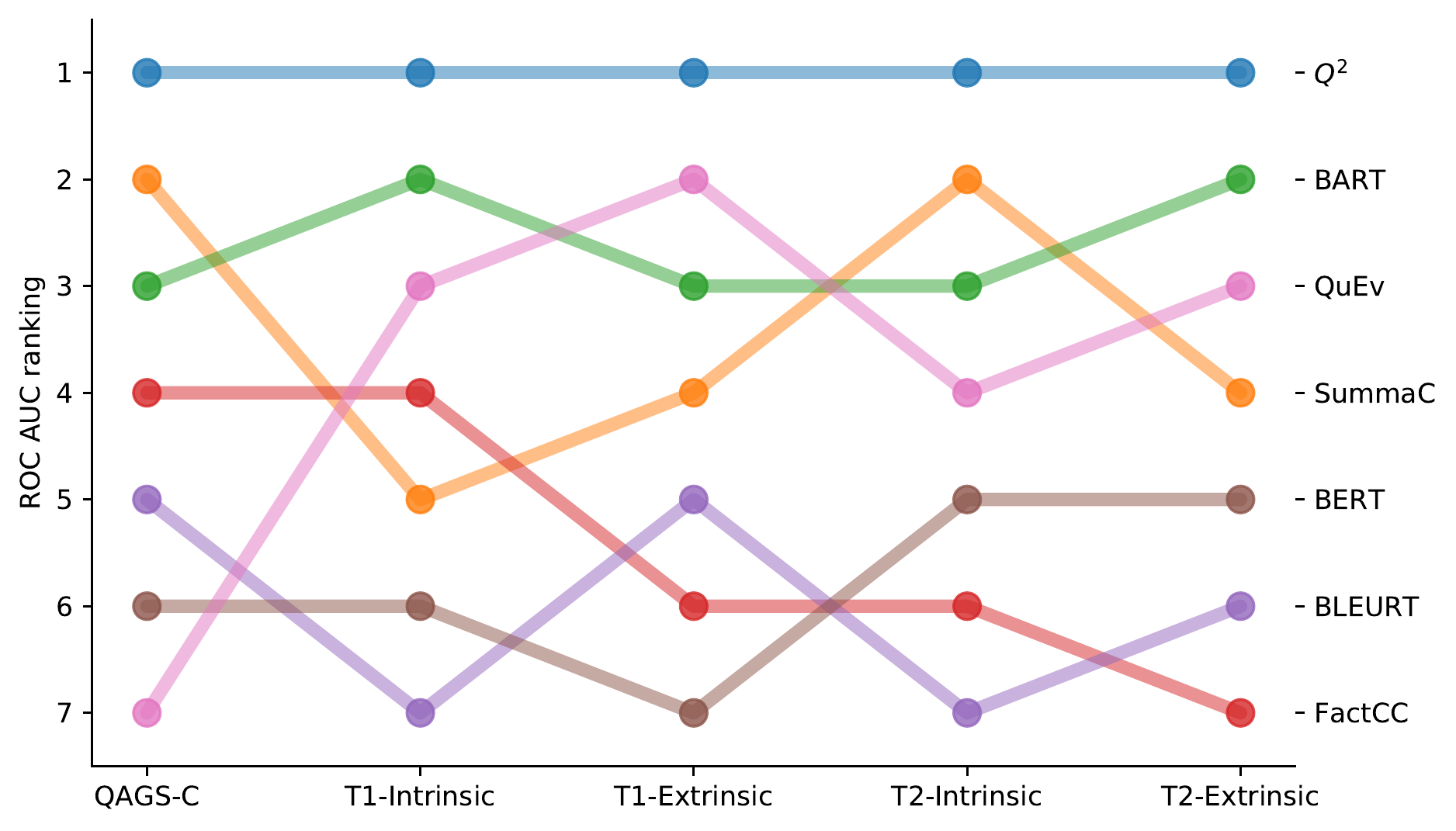}
    \caption{
        {\bf ROC AUC ranking of faithfulness metrics under different datasets}. Only faithfulness metrics used in both \citep{honovich-etal-2022-true} and Table~\ref{tab:auc} are presented. T1: Task 1 of \dataset, T2: Task 2 of \dataset, Intrinsic: Intrinsic Errors, Extrinsic: Extrinsic Errors, BART: \textsc{BARTScore}, BERT: \textsc{BERTScore}, QuEv: \textsc{QuestEval}. 
    }
    \label{fig:ranking}
\end{figure}

Furthermore, we compare our ROC AUC scores with those in existing datasets as shown in TRUE \citep{honovich-etal-2022-true}. In \dataset, faithful and unfaithful samples under each error type are balanced for both Tasks~1 and 2. Therefore, for a fair comparison, we pick QAGS-C \citep{wang-etal-2020-asking} (also a balanced dataset on CNN/DailyMail) in TRUE.
In Table~\ref{tab:auc}, it shows that
the ROC AUC scores evaluated on \dataset are generally much smaller (50--70$\%$ with many values close to random baseline),
whereas most ROC AUC scores are 70--84$\%$ in QAGS-C (see Appendix~\ref{sec:ROC_AUC_other_paper}).
This again indicates that the human-generated errors in \dataset are more difficult for metrics to detect than model-generated errors in existing datasets, reinforcing the value of \dataset as a challenging benchmark for evaluating faithfulness metrics.
% \joel{one thing we need to be wary of is someone getting the wrong idea about our paper thinking that it is about detecting unfaithful human summaries, and then ask when would this happen?  We need to be clear that we are using human to make errors because we believe it will make for a more difficult and useful evaluation and also we can track the problematic error areas.}
In addition, we also compare the ROC AUC rankings of different faithfulness metrics under QAGS-C and \dataset. Specifically, we summarize the performance rankings under QAGS-C from Appendix~\ref{sec:ROC_AUC_other_paper} as well as those from Table~\ref{tab:auc} under Intrinsic/Extrinsic Error types in Tasks~1 and 2, and report them in Figure~\ref{fig:ranking}, where only faithfulness metrics used in both \citep{honovich-etal-2022-true} and Table~\ref{tab:auc} are presented.
In Figure~\ref{fig:ranking}, we observe that for some faithfulness metrics, such as \textsc{$Q^2$} and \textsc{BARTScore}, their ROC AUC rankings are quite stable across all datasets. However, for other faithfulness metrics, the performance ranking under QAGS-C
%\textsc{$Q^2$ > SummaC >  BARTScore > FactCC > BLEURT > BERTScore > QuestEval} (see Appendix~\ref{sec:ROC_AUC_other_paper}), which 
is very different from the ranking derived from \dataset, e.g., \textsc{QuestEval} mostly exhibits high ROC AUC ranking in \dataset; by contrast, it experiences the worst performance in QAGS-C. 
% Therefore, it is critical to thoroughly evaluate the faithfulness metrics using our human-generated \dataset to better understand their performance. 
Thus, we believe \dataset complements existing benchmarks and allows a more comprehensive analysis of faithfulness metrics in future studies.
%\rob{Comparisons to existing benchmarks require grounding. For example a comparison of the overall scores to TRUE should be in a table. If we don't have space it should at least be placed and referenced in the abstract.}

\paragraph{Qualitative Analysis.}

We provide a qualitative analysis of examples that demonstrate the increased difficulty of Task~2. % are more challenging than the edits in Task~1.
%and 2) how edited unfaithful summaries with incorrect error types in Task~1 affect our final error type distributions after error type relabeling.
%from each bucket within Task~1 and 2 and list them in Table \ref{tab:casestudy1} and Table \ref{tab:casestudy2}, respectively.
%\lm{Do we have examples to show that unfaithful summaries in Task 2 are harder than Task 1?}
% \paragraph{Detection Difficulty.}
% For Task~1, two examples from the edited summaries are identified to study the unfaithfulness detection difficulties (first two examples in Table \ref{tab:casestudy1}). 
% Both of these edited summaries satisfy the required error types. However, for the introduced extrinsic entity error in the first example, it is an easy case generally for all metrics, which is also evidenced by our observations in Table~\ref{tab:sensitivity1} that extrinsic entity errors are the easiest in Task~1. In this example, all metrics succeed to give a higher faithful score to the reference summary.
% By comparison, in the second example, the edited summary contains a predicate error; furthermore, the edited word also appears in the original article (i.e., intrinsic error, which is more challenging according to \ref{sec:eval_result}), which therefore constitutes an unfaithful summary that almost all metrics struggle with. In this example, only 4 metrics manage to give a higher faithful score to the reference summary (i.e., CoCo, DAE, FactCC, and BART). 
% For Task~2, similar to Task~1, we also identify two unfaithful edited summaries in Table~\ref{tab:casestudy2}, both with extrinsic entity errors. 
The examples are provided in Table~\ref{tab:casestudy}.
Each row contains edited summaries from Tasks~1 and 2 for the same original article and its reference summary. In addition, to compare edited summaries under the same error type, we pick examples where the corrected error type from Task~1 is the same as the exhibited error type from Task~2.
As shown in Table~\ref{tab:casestudy}, in the first example, for the \textit{Extrinsic Entity Error} type, the annotator in Task~1 modifies the entity \textit{elevator shaft} to another entity \textit{staircase}. Whereas the annotator in Task~2 modifies the word \textit{communal} to \textit{private} (i.e., also an \textit{Extrinsic Entity Error}) which requires commonsense knowledge to infer that \textit{private} is contradictory to the fact that \textit{the elevator is used by several businesses in the building}. In the second example, for the \textit{Extrinsic Entity Error} type, the annotator in Task~1 modifies the entity name from \textit{Claire Nugent} to a random name \textit{Tim Horton}, whereas the annotator in Task~2 changes \textit{record player} to \textit{black and white TV} to fit the \textit{60s} theme, which again, requires additional knowledge. In the last example, the annotator in Task~2 modifies the temporal state of the action \textit{sign} from \textit{signed} to \textit{is considering signing} which is more challenging than changing the action \textit{urged} to its antonym \textit{discouraged} as the annotator in Task~1 does.

For the first two examples of Task~2 in Table~\ref{tab:casestudy}, only 4 metrics (\textsc{QAFactEval}, \textsc{QuestEval}, \textsc{BLEURT}, and \textsc{BERTScore} for the first example; \textsc{QAFactEval}, \textsc{SummaC}, \textsc{BLEURT}, and \textsc{ROUGE-2} for the second example) succeed in giving a higher score to the reference summary. In comparison, 9 and 11 metrics succeed in giving a higher score to the reference summary in their Task~1 counterparts, respectively. For the last example, 8 metrics succeed in Task~2 and all 12 metrics succeed in Task~1. Thus, Table~\ref{tab:casestudy} shows that some unfaithful summaries in Task~2 are more challenging for faithfulness metrics to detect, which further exemplifies the challenges of Task~2 in \dataset. 
%are able to successfully fool the evaluation metrics by landing a high faithful score from different metrics ().

%pattern corresponds to the samples where a perfect unfaithful summary is introduced to the edited summary such as Example~1. In the second pattern (Example~2), the annotators regard the quantity of a \emph{Noun object} as a circumstance and make edits to the quantity, hence mistakenly regard Entity Errors as Circumstance Errors. Similarly, in some cases (Example~3), the annotators mistake Intrinsic Errors as Extrinsic Errors by failing to identify that the edits are mentioned in the original article. Both of these errors are corrected in the relabling process of Task~1.
%The fourth category (Example~4) represents the cases where making unfaithful edits of a certain error type is more challenging, leading to annotators being unable to provide an edit. 

%Similar to Task~1, the first pattern corresponds to the samples where a perfectly edited unfaithful error is introduced as shown in Example~1 in Table \ref{tab:casestudy2}. In the second pattern (Example~2), the edits are able to fool the metrics by predicting a high faithful score from different metrics (QAFactEval in this example, the faithful score is $0.899$). 
% \lm{if there is no unfaithful error in the last row, let's remove it}
%In the last category (Example~3), the edits are not considered unfaithful or erroneous. 

\section{Conclusion}
In this paper, we presented a benchmark of unfaithful minimal pairs (\dataset) to evaluate faithfulness metrics.
Unlike prior work where all unfaithful summaries are model generated, each unfaithful summary in \dataset is generated by minimal human edits to introduce one unfaithful error given a reference summary.
% We showed that unfaithful summaries in \dataset are substantially different from those in other datasets, and more challenging to be identified. 
% We analyzed faithfulness metrics according to seven error types in our proposed taxonomy via consistency and ROC AUC protocols.
% Evaluation results show that no metrics simultaneously achieve the best performance in these two protocols. 
% Nevertheless, \textit{Intrinsic Error}s are consistently more difficult than extrinsic ones for all metrics. 
% %Furthermore, the metric performance ranking under \dataset is significantly different from that using model generated datasets. 
% Therefore, \dataset provides a valuable dataset that complements the existing resources in testing faithfulness evaluation metrics.
% % \joel{could we even say that the community should use this one as opposed to the other ones?  Basically, is there a reason to use the others at this point?}
% % \rob{a) I don't think so (human-generated isn't necc. a setting ppl care about, plus sample sizes are much larger for other datasets), and b) authors of the other datasets are likely going to review this paper, framing \dataset as a replacement instead of a supplement will probably make them more defensive :)}
Through our experiments, we found that \dataset complements existing benchmarks in a number of ways.
First, we found that the summaries in \dataset are harder to discriminate and less probable under SOTA summarization models.
Second, we used \dataset to measure the consistency of metrics, which cannot be readily measured using other benchmarks.
This analysis revealed a discrepancy between the discriminability and consistency of existing metrics, highlighting an important area for future faithfulness metric research to address.
Finally, we used \dataset to study faithfulness metrics' performance on individual error types---where our minimal-pair-inspired setup helped control for conclusions being conflated across multiple error types---which revealed that sensitivity to intrinsic errors is another important area for future research to focus on.

% \joel{I think the conclusion is a nice summary of the work but could sell the work much better. For example, should practitioners use our dataset over the prior ones? Do the difference in rankings from ours and prior art show something fundamentally wrong in our field?  I just don't want it to end with here's another dataset you could use and leave it at that.}

% \joel{In fact the last para of the Introduction and the last para of Related Work have really nice specific details on why we are better than others.  Those points are not reinforced in the conclusion.}

%\clearpage
\section*{Acknowledgements}
We would like to thank our colleague Aoife Cahill at Dataminr for her valuable comments, suggestions, and support for this paper.
We also thank the anonymous reviewers for their feedback and comments.

\section*{Limitations}

% While \dataset contains 889 human-written unfaithful summaries, we only ask one annotator to create one unfaithful summary for each document-summary pair and a given error type.
% We acknowledge that having more unfaithful summaries created by different annotators can improve the summary diversity for better evaluating faithfulness metrics, but unfortunately that is beyond our currently available resources.
Although \dataset is, to our knowledge, the first dataset on which to study the consistency of faithfulness metrics on human-written errors across different error types, there are some limitations regarding the conclusions that can be drawn from it.
For one, because \dataset is comprised of minimal edits to reference summaries from CNN/DailyMail, it is not suitable for analyzing the consistency of faithfulness metrics when errors are added to reference summaries already containing many errors.
In addition, due to a combination of resource constraints and human preferences for writing specific types of errors, the sample sizes for some error types in Task~2 (e.g., \textit{Coreference Error} and \textit{Intrinsic Predicate Error}) may not be sufficiently large to enable statistically significant comparisons between different metrics for specific error types.

\section*{Ethics Statement}

The collection of \dataset involves human annotations. 
The human annotators are provided with clear task instructions and informed of the conditions where they would be qualified and disqualified.
% For compensation, 
We compensate annotators with $\$3.00$ per assignment in the qualification task and $\$0.50$ per assignment in the full task for both Tasks 1 and 2. The final paid rate is $\$15$ per hour which is over the US national minimum wage\footnote{\url{https://www.dol.gov/general/topic/wages/minimumwage}} of $\$7.25$. 
% \ke{add minimum hourly wage; the data might be misused to train an unfaithful summarization model.}
We are also aware that our shared datasets could be potentially misused as training samples, albeit a small number, to develop models to generate unfaithful content. 

\bibliography{custom}
\bibliographystyle{acl_natbib}

%\clearpage
\appendix

\renewcommand{\thefigure}{A\arabic{figure}}
\renewcommand{\thetable}{A\arabic{table}}

\setcounter{figure}{0}
\setcounter{table}{0}

\section{Details of Task~1: Taxonomy-based Unfaithful Summaries}
\label{sec:appendix1}

\subsection{Qualification Task}
\begin{figure*}
  \includegraphics[width=\linewidth]{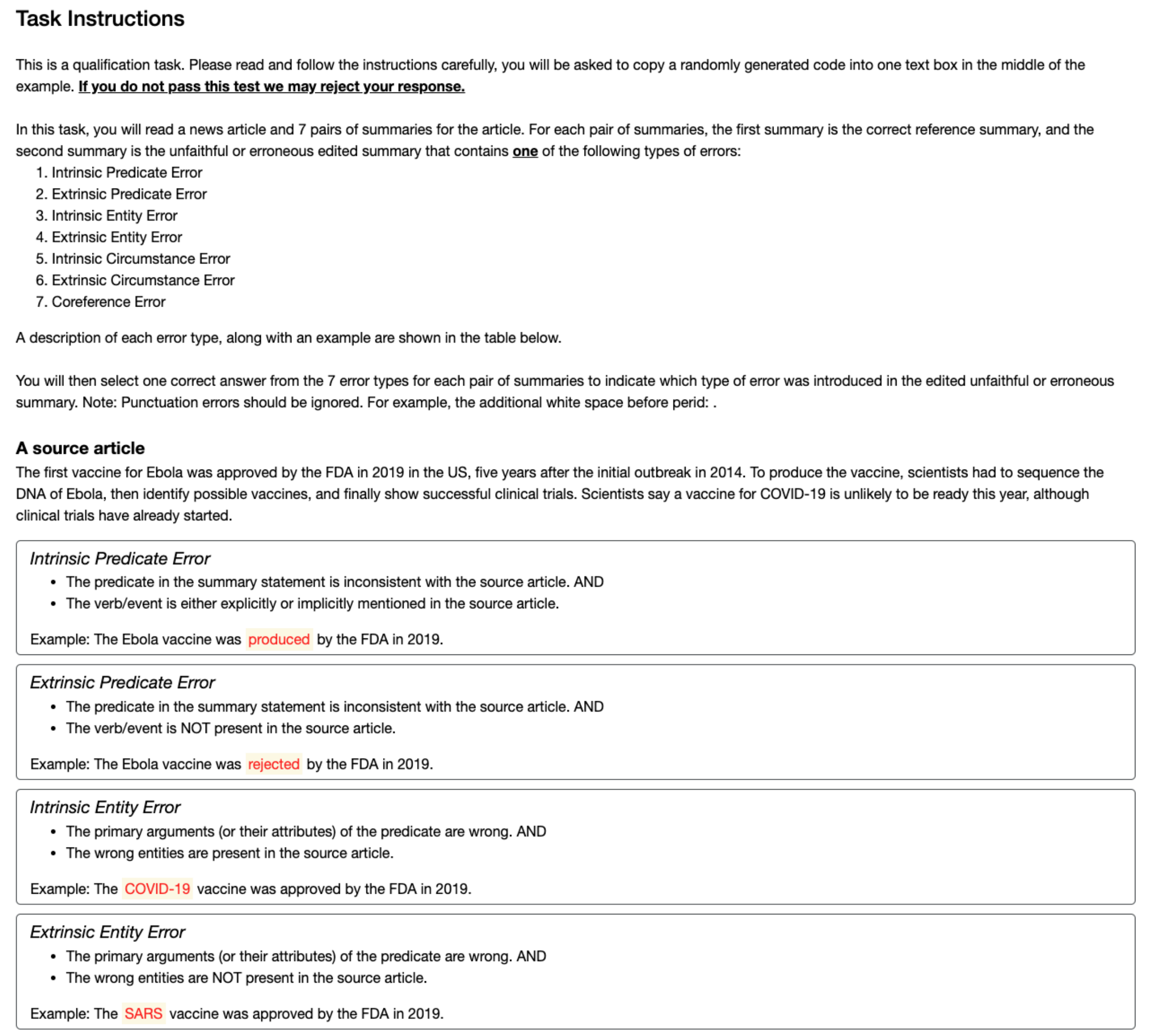}
  \caption{Screenshot of the qualification task for Task~1 (1/4).}
  \label{fig:Task_1_qual_1}
\end{figure*}

\begin{figure*}
  \includegraphics[width=\linewidth]{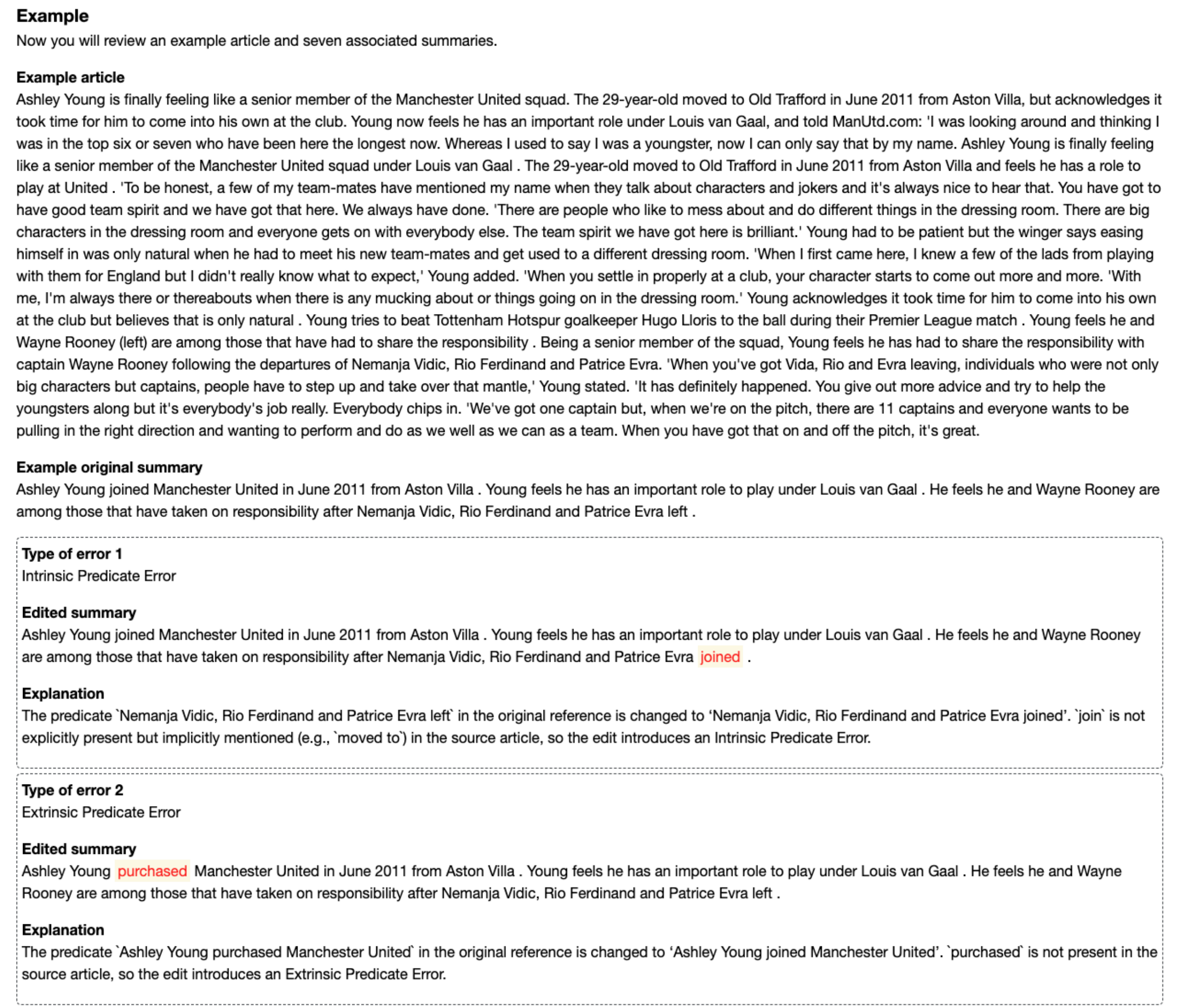}
  \caption{Screenshot of the qualification task for Task~1 (2/4).}
  \label{fig:Task_1_qual_2}
\end{figure*}

\begin{figure*}
  \includegraphics[width=\linewidth]{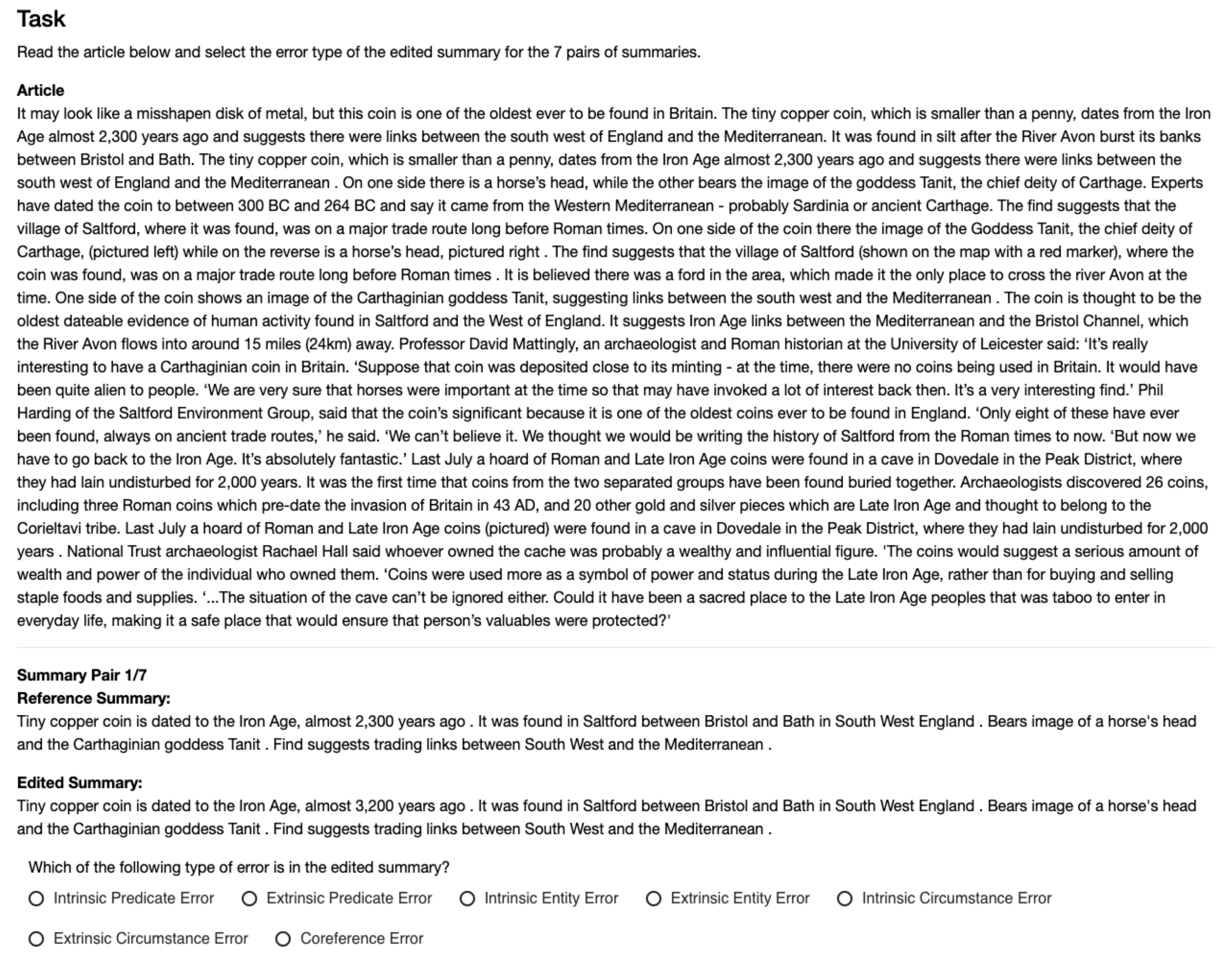}
  \caption{Screenshot of the qualification task for Task~1 (3/4).}
  \label{fig:Task_1_qual_3}
\end{figure*}

\begin{figure*}
  \includegraphics[width=\linewidth]{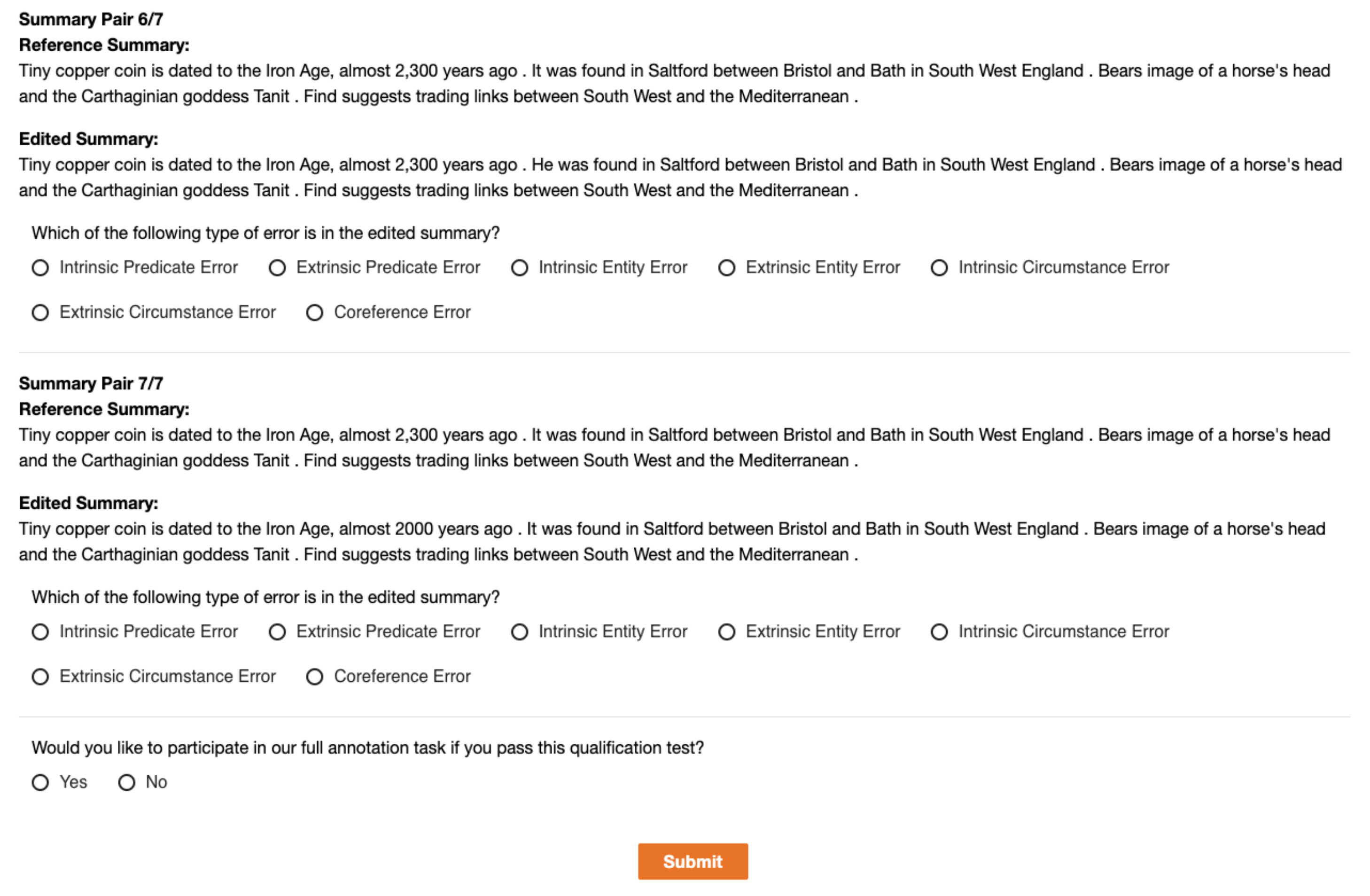}
  \caption{Screenshot of the qualification task for Task~1 (4/4).}
  \label{fig:Task_1_qual_4}
\end{figure*}

The instructions and the task interface for the qualification task of Task~1 are shown in Figures~\ref{fig:Task_1_qual_1} to \ref{fig:Task_1_qual_4}. 

In this qualification task, all US-based annotators are able to participate. Specifically, we ask annotators to read a news article and seven pairs of summaries. For each pair of summaries, the first summary is the correct reference summary, and the second summary is the unfaithfully edited summary that contains one of the seven error types in our taxonomy. We then ask the annotators to select one answer from the seven error types to indicate which type of error is introduced in the edited unfaithful summary. Only the annotators who answered 6 out of these 7 questions correctly passed the qualification task. We launched 3 batches in total with 9 assignments for each batch, and 9 annotators passed the qualification task.

\subsection{Full Task}
\begin{figure*}
  \includegraphics[width=\linewidth]{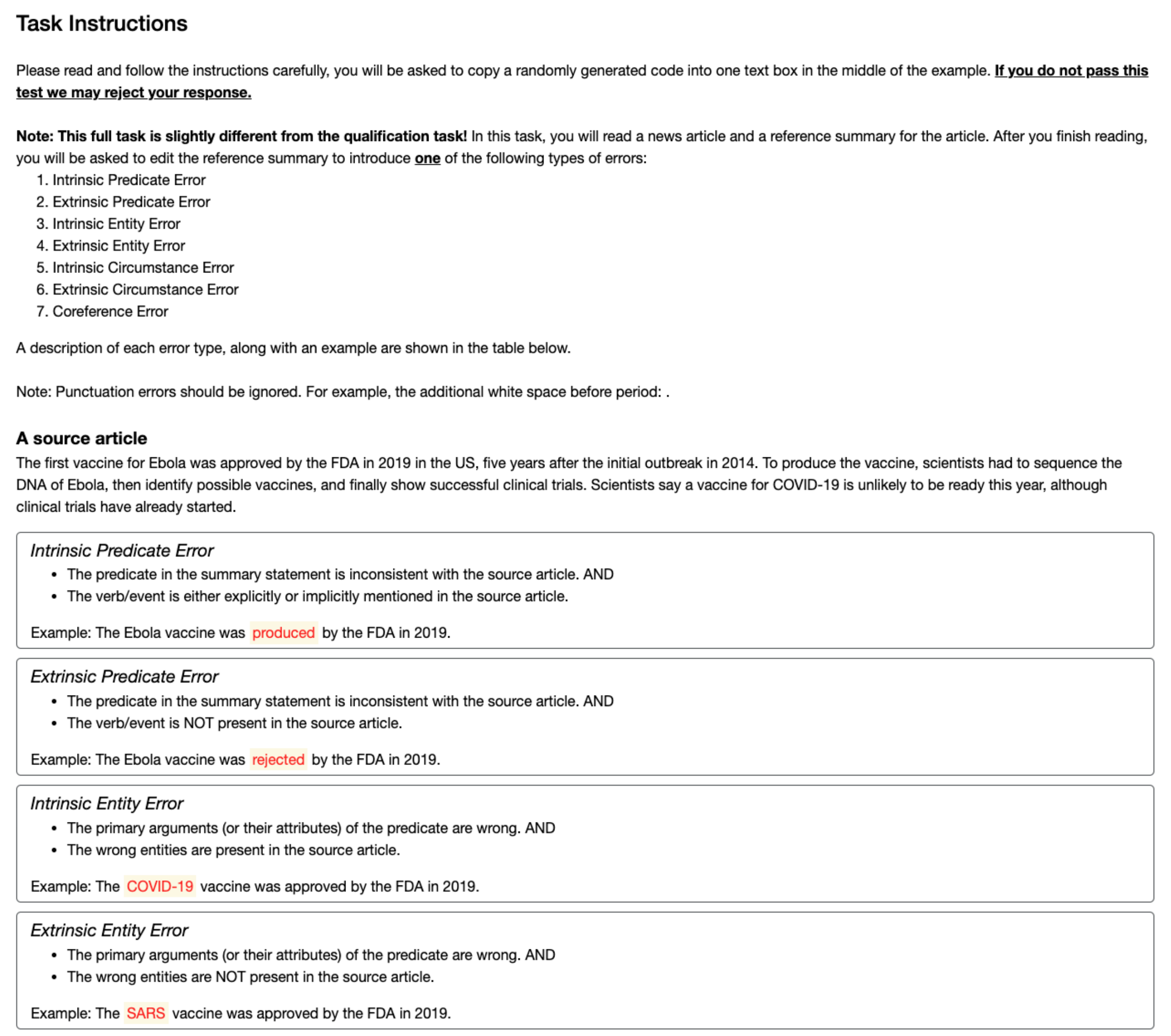}
  \caption{Screenshot of the full task for Task~1 (1/3).}
  \label{fig:Task_1_full_1}
\end{figure*}

\begin{figure*}
  \includegraphics[width=\linewidth]{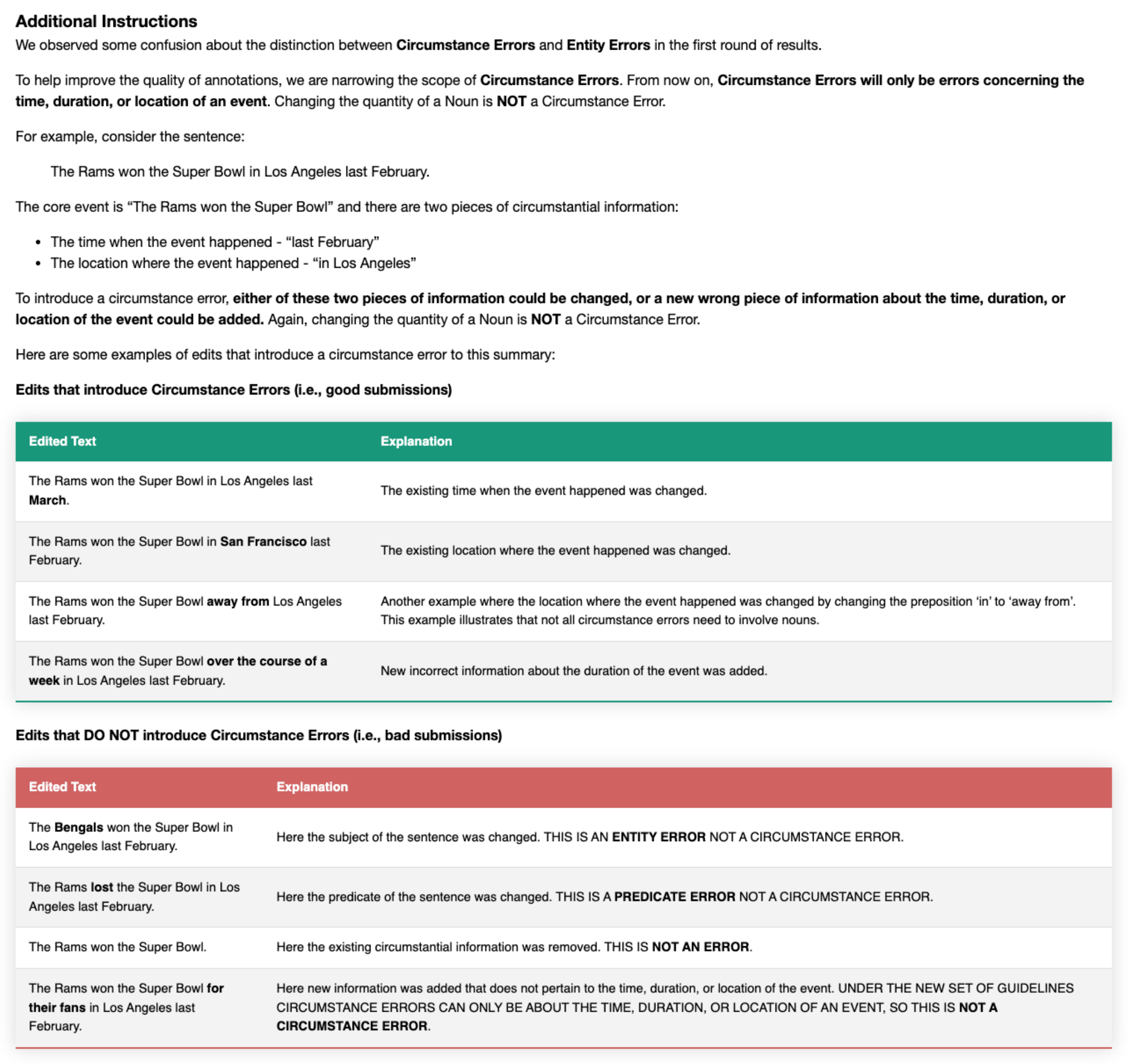}
  \caption{Screenshot of the full task for Task~1 (2/3).}
  \label{fig:Task_1_full_2}
\end{figure*}

\begin{figure*}
  \includegraphics[width=\linewidth]{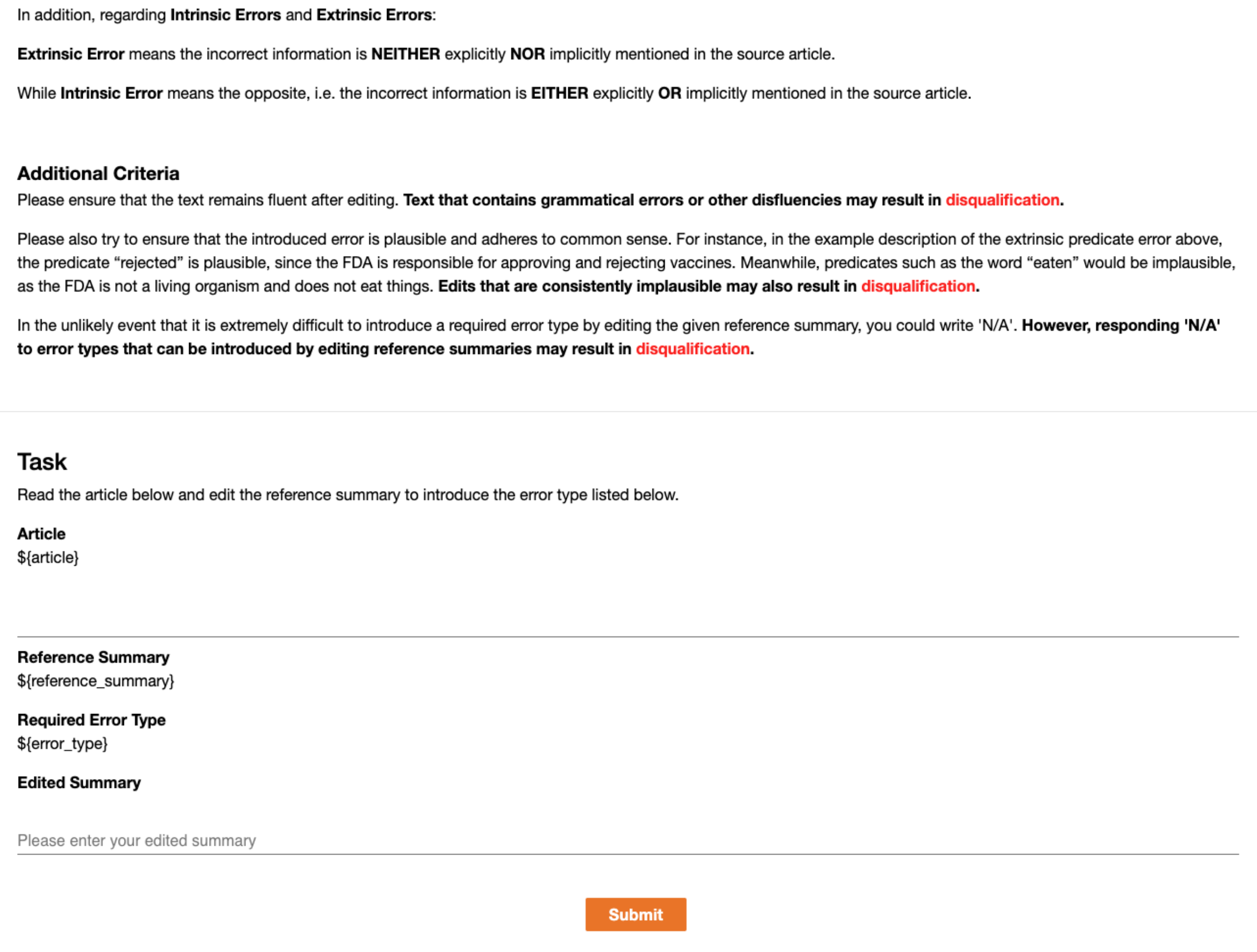}
  \caption{Screenshot of the full task for Task~1 (3/3).}
  \label{fig:Task_1_full_3}
\end{figure*}

The instructions and the task interface for the full task of Task~1 are shown in Figures~\ref{fig:Task_1_full_1} to \ref{fig:Task_1_full_3}.

In the full task for Task~1, different from the qualification task, we ask the annotators to read a news article from CNN/DailyMail \citep{hermann2015teaching} and one reference summary for the article. We then ask the annotators to edit the reference summary to introduce the error type specified through a minimal edit. If they cannot introduce the error type based on the reference summary, they can write ``N/A'' to indicate that it is impossible to introduce the specified error type based on the provided reference summary. There are 18 samples in Task~1 dataset that are annotated as ``N/A'' by the annotators, all of which are reviewed by the authors of this paper and re-annotated with the correct edits (we note that the required error types can be provided for all these cases) as a post-processing step to ensure the completeness of the dataset. 

In addition, for Task~1, to help reduce the confusion from annotators regarding \textit{Circumstance Error}s and \textit{Entity Error}s, we explicitly specify that the \textit{Circumstance Error}s should only be erroneous edits concerning the time, duration, or location of an event, and changing the quantity of a \emph{noun} is not considered as a \textit{Circumstance Error}.

\section{Details of Task 2: Freestyle Unfaithful Summaries}
\label{sec:appendix2}

\subsection{Qualification Task}
\begin{figure*}
  \includegraphics[width=\linewidth]{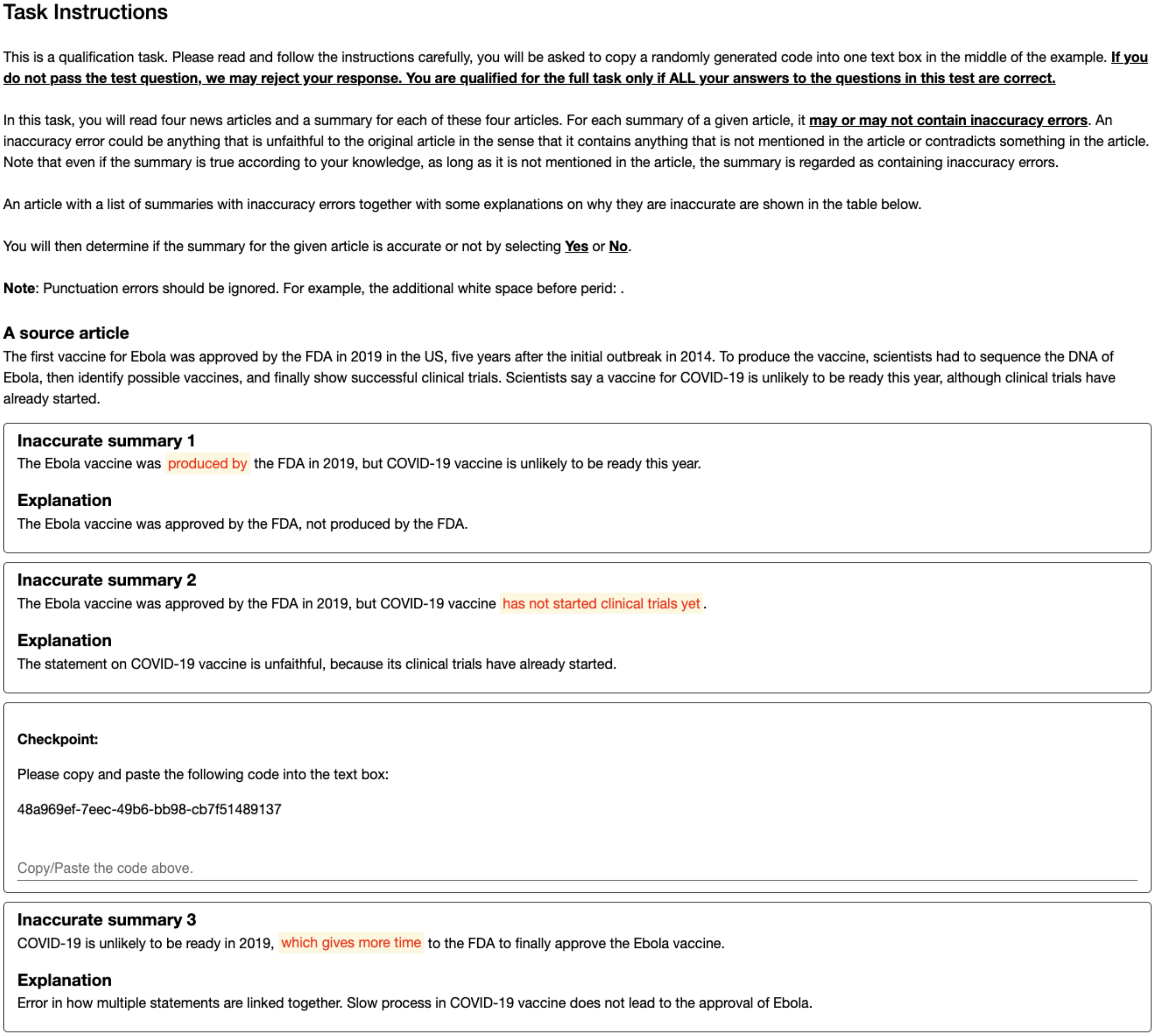}
  \caption{Screenshot of the qualification task for Task 2 (1/2).}
  \label{fig:Task_2_qual_1}
\end{figure*}

\begin{figure*}
  \includegraphics[width=\linewidth]{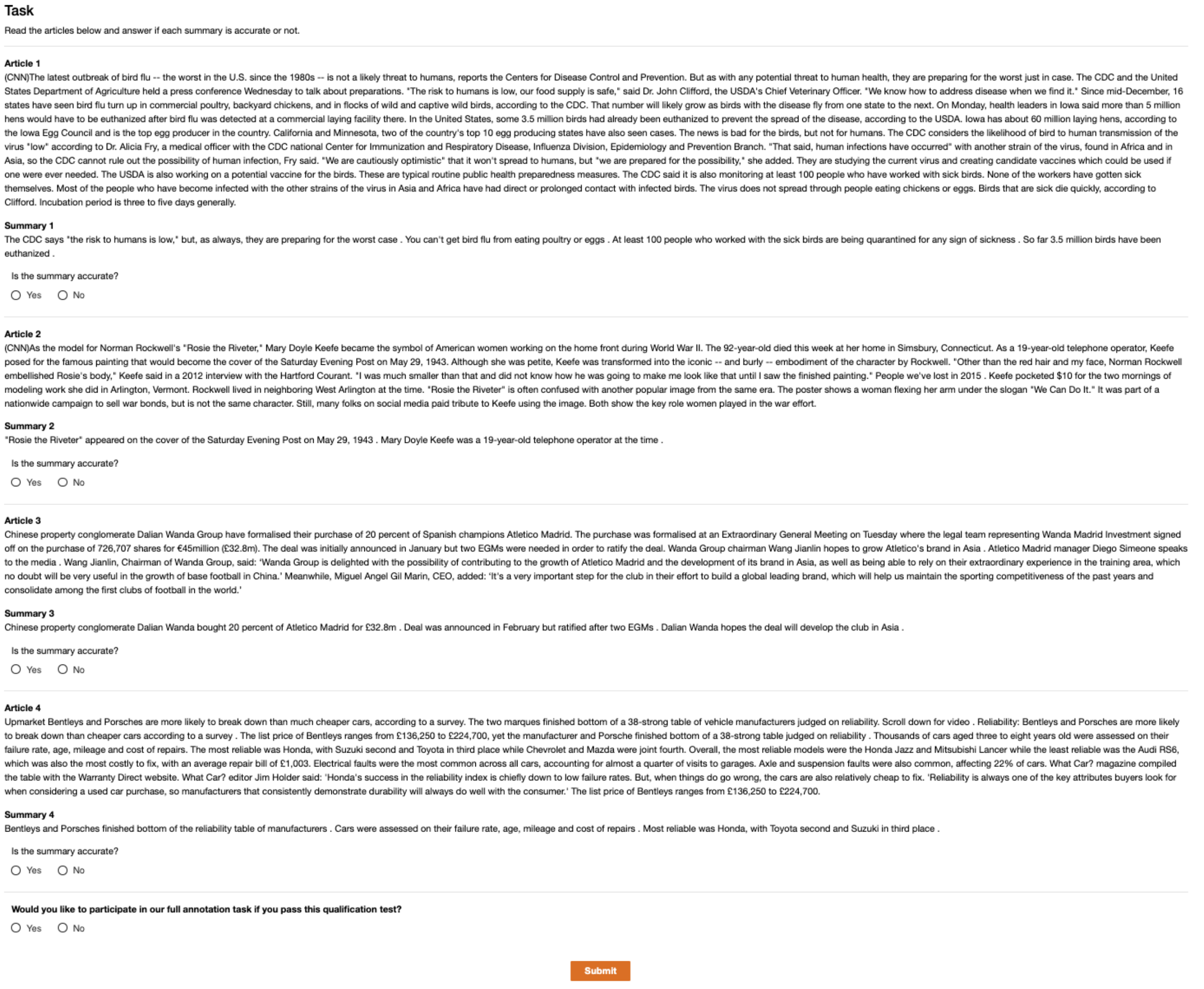}
  \caption{Screenshot of the qualification task for Task 2 (2/2).}
  \label{fig:Task_2_qual_2}
\end{figure*}

The instructions and the task interface for the qualification task of Task~2 are shown in Figures~\ref{fig:Task_2_qual_1} to \ref{fig:Task_2_qual_2}. 

In this qualification task, all US-based annotators who did not participate in the qualification task of Task~1 are qualified to participate. Specifically, we show the annotators four pairs of news article and its summary from CNN/DailyMail, and ask them to answer if the summaries are faithful based on the original news articles. Among the four pairs, three of them are unfaithful and one is faithful. Only the annotators who answered correctly to all of these 4 pairs passed the qualification task. We launched 3 batches in total with 9 assignments for each batch, and 8 annotators passed the qualification task.

\subsection{Full Task}
\begin{figure*}
  \includegraphics[width=\linewidth]{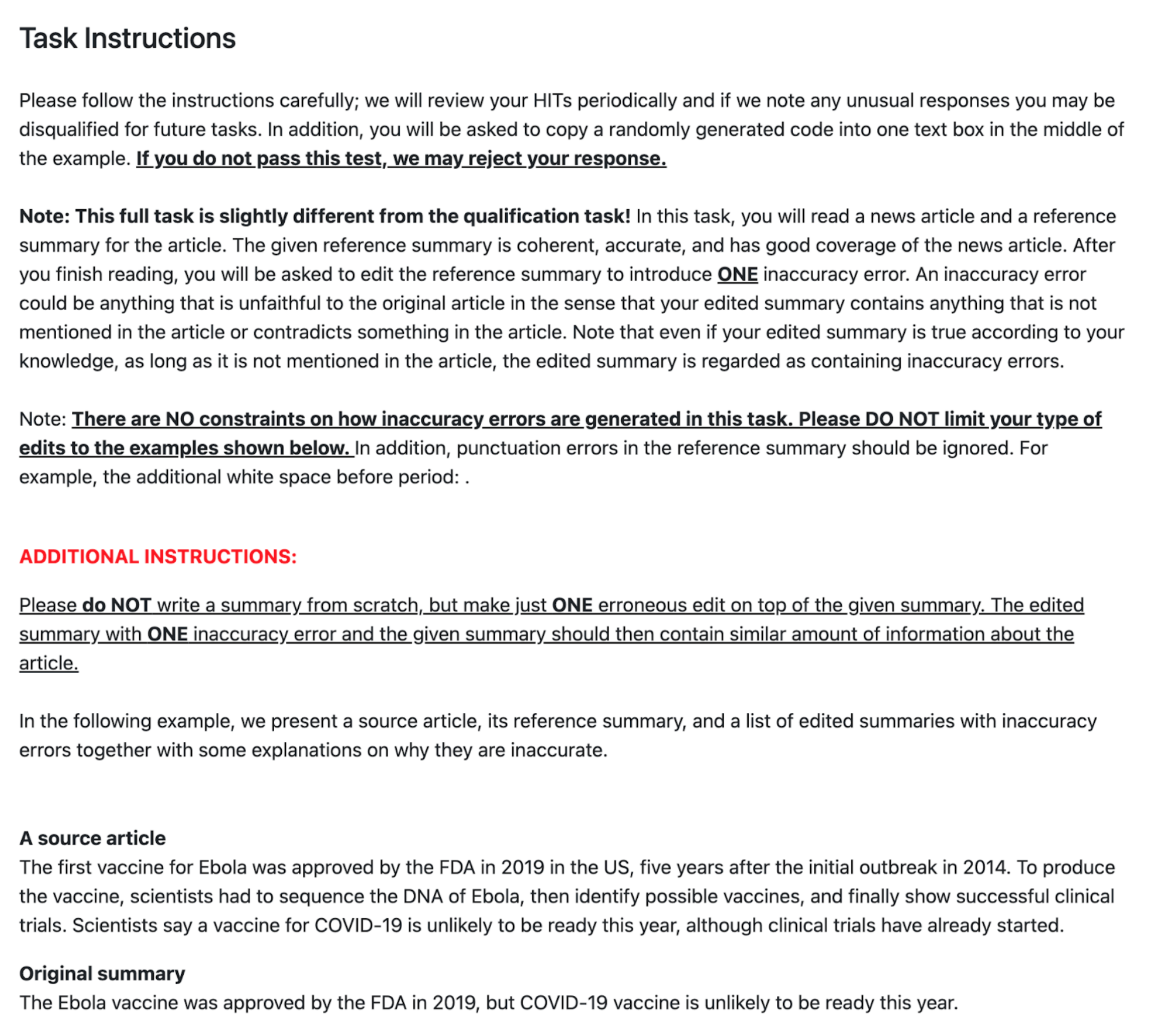}
  \caption{Screenshot of the full task for Task 2 (1/2).}
  \label{fig:Task_2_full_1}
\end{figure*}

\begin{figure*}
  \includegraphics[width=\linewidth]{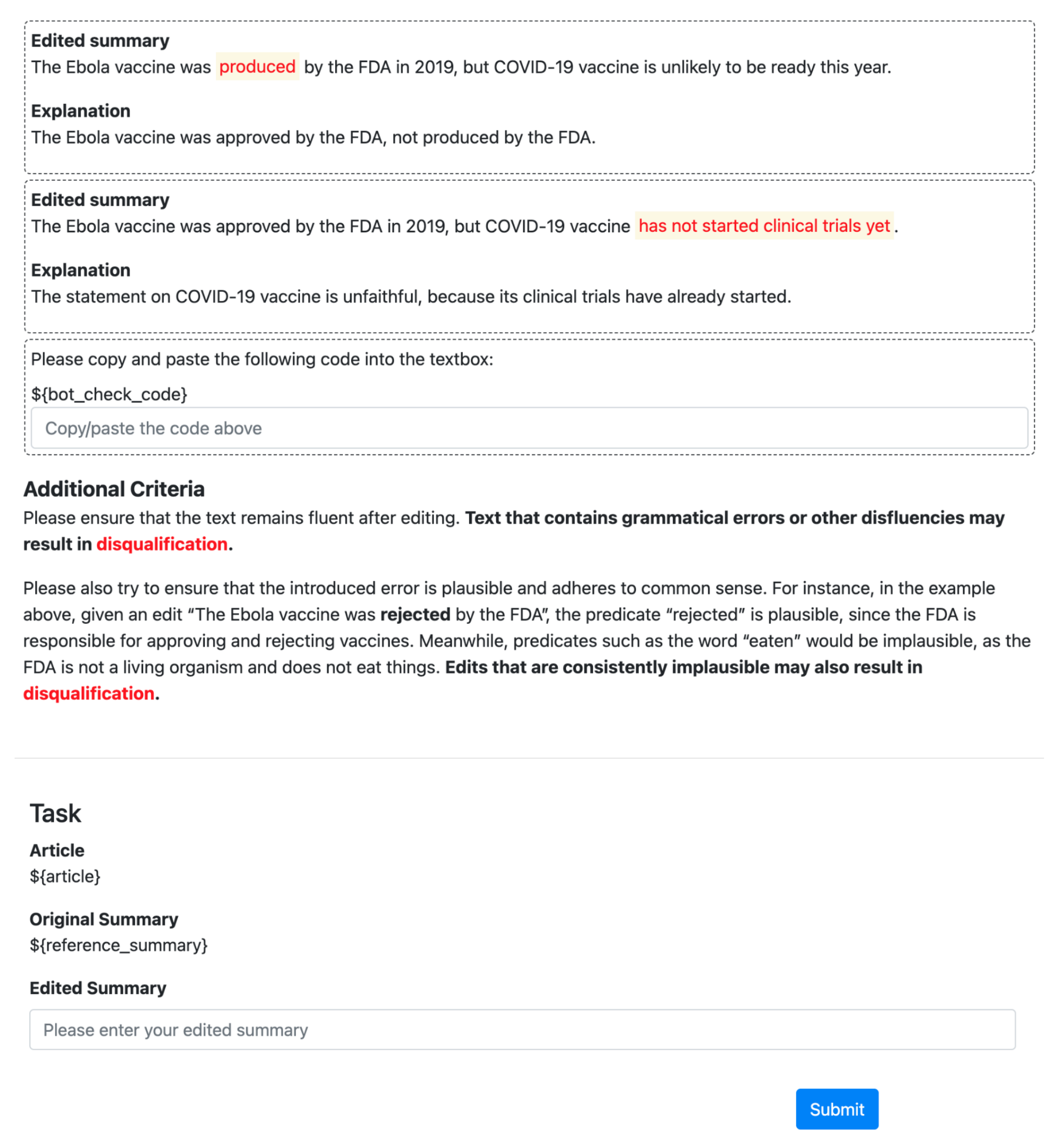}
  \caption{Screenshot of the full task for Task 2 (2/2).}
  \label{fig:Task_2_full_2}
\end{figure*}

The instructions and the task interface for the full task of Task~2 are shown in Figures~\ref{fig:Task_2_full_1} to \ref{fig:Task_2_full_2}. 

In the full task of Task~2, unlike Task~1, we do not list any potential error types so as to achieve freestyle editing. The edited summary is valid as long as only one error is introduced based on the reference summary via a minimal edit. Furthermore, we also do the following to ensure the quality of edited summaries:
\begin{itemize}
    \item For minimal edits, we explicitly ask annotators \emph{not} to write from scratch, but to introduce only one error on top of the given reference summary.
    \item In the pilot study, we notice that some edited summaries are simply removing/adding sentences or phrases (such data points are removed in the final released data); we, therefore, add additional instructions that require the edited and the reference summaries to contain a similar amount of information about the given news article (i.e., similar coverage).
    \item The edited summaries should be grammatically correct.
    \item The edited summaries should be plausible and adhere to common sense.
    \item Some examples of edited summaries are given in the task instructions.
\end{itemize}

\section{ROC AUC Results from Other Benchmarks}
\label{sec:ROC_AUC_other_paper}
To compare \dataset with other benchmarks, we also report the ROC AUC scores from TRUE \citep{honovich-etal-2022-true}. Specifically, in \dataset, faithful and 
unfaithful samples under each error type are balanced for both Tasks 1 and 2. Therefore, for a fair comparison, 1) in TRUE, we pick QAGS-C \citep{wang-etal-2020-asking}, which is also a balanced dataset on CNN/DailyMail. The ROC AUC scores of QAGS-C are reported in Table~\ref{tab:TRUE_results}; 2) for faithfulness metrics in Table~\ref{tab:TRUE_results}, we use the same implementation and model checkpoints in this paper as those in TRUE \citep{honovich-etal-2022-true}. Then according to Table~\ref{tab:TRUE_results}, the metric performance ranking in terms of ROC AUC for QAGS-C is \textsc{$Q^2$ > SummaC >  BARTScore > FactCC > BLEURT > BERTScore > QuestEval}, 
which is very different from the ranking derived from our \dataset dataset, e.g., \textsc{SummaC} exhibits worse ROC AUC than \textsc{QuestEval} for most error types in both Tasks~1 and 2 (see Table~\ref{tab:auc}).

In addition to the balanced dataset QAGS-C in TRUE, we also report the ROC AUC scores of imbalanced FRANK \citep{pagnoni-etal-2021-understanding} and SummEval \citep{fabbri-etal-2021-summeval} datasets (two datasets containing CNN/DailyMail) from TRUE in Table~\ref{tab:TRUE_results}. Although the FRANK and SummEval datasets are imbalanced, we have similar observations as those from the QAGS-C dataset: 1) their ROC AUC scores (mostly 70--90$\%$) are much larger than the ROC AUC scores (50--70$\%$) derived from our \dataset dataset; 2) in terms of the ROC AUC ranking, the top two remain \textsc{$Q^2$} and \textsc{SummaC} for both FRANK and SummEval, and \textsc{SummaC} always ranks higher than \textsc{QuestEval}. By contrast, in Table~\ref{tab:auc}, we show that \textsc{SummaC} mostly exhibits worse ROC AUC than \textsc{QuestEval}.

%\paragraph{Remark.} In \dataset, the faithful and unfaithful samples are balanced. For a fair comparison with ROC AUC results in TRUE \citep{honovich-etal-2022-true}, the dataset we use from TRUE should also be balanced. From Table~2 in TRUE \citep{honovich-etal-2022-true}, we see that the total numbers of faithful and unfaithful samples (when FRANK \citep{pagnoni-etal-2021-understanding}, SummEval \citep{fabbri-etal-2021-summeval} and QAGS-C \citep{wang-etal-2020-asking} are combined) are $1,642$ and $864$, respectively. To make this dataset balanced, we can over-sample unfaithful samples by duplicating all unfaithful samples (i.e., each unfaithful sample appears twice in the over-sampled dataset). Then under such over-sampled dataset, the False Positive and True Positive rates remain the same comparing to the dataset without over-sampling under the same decision threshold, thus not affecting the ROC AUC scores. Therefore, the scores in Table~\ref{tab:TRUE_results} remain the same when unfaithful samples are over-sampled as above.

\begin{table*}[b]
    \small
    \centering
    \begin{tabular}{lcccccccccccc}
        \toprule
	&	\textsc{$Q^2$}	&	\textsc{SummaC}	&	\textsc{BARTscore}	&	\textsc{FactCC}	&	\textsc{BLEURT}	&	\textsc{BERTscore}	&	\textsc{QuestEval}	\\
 \midrule
QAGS-C	&	83.5	&	80.9	&	80.9	&	76.4	&	71.6	&	69.1	&	64.2	\\
\midrule
FRANK	&	87.8	&	89.1	&	86.1	&	76.4	&	82.8	&	84.3	&	84.0	\\
SummEval	&	78.8	&	81.7	&	73.5	&	75.9	&	66.7	&	77.2	&	70.1	\\
         \bottomrule
    \end{tabular}  
    \caption{\textbf{ROC AUC (\%) of faithfulness evaluation metrics in TRUE} \citep{honovich-etal-2022-true}. All datasets contain CNN/DailyMail. Faithful and unfaithful samples in QAGS-C are balanced; however, in FRANK and SummEval, faithful and unfaithful samples are imbalanced. 
    }
    \label{tab:TRUE_results}
\end{table*}

%For the data post-processing of this task, we (authors of this paper) manually assign an error type to each unfaithful summary according to our error type taxonomy in Task~1. We acknowledge that there may be cases where the induced unfaithful error belongs to more than one error type or cannot be categorized by any error types. When this happens, we assign it to type ``Other''. Our final results show that the rate to have label ``Other'' is only $2.5\%$, indicating that our error type taxonomy is able to cover most cases. 

\end{document}